\documentclass[sn-mathphys,Numbered]{sn-jnl}

\usepackage{graphicx}%
\usepackage{amsmath,amssymb,amsfonts}%
\usepackage[title]{appendix}%
\usepackage{textcomp}%
\usepackage{manyfoot}%
\usepackage{booktabs}%

\usepackage{mathtools}
\usepackage{accents} 
\usepackage{lineno}
\usepackage{cleveref}
\usepackage{hyperref}
\usepackage{relsize}
\usepackage[nice]{nicefrac}
\usepackage{bigints}
\usepackage{derivative}  
\usepackage[table]{xcolor}
\newcommand{\defeq}{\vcentcolon=}

\newcommand{\Rbb}{\ensuremath{\mathbb{R}}} 

\newcommand{\bcheck}[1]{\ensuremath{\accentset{\vee}{#1}}}
\newcommand{\ho}[1]{\ensuremath{\mathring{#1}}}
\newcommand{\abs}[1]{\left| #1 \right|} 
\newcommand{\avg}[1]{\left< #1 \right>} 
\newcommand{\de}[1]{\mathrm{d} #1} 

\DeclareMathOperator\erfc{erfc}
\begin{document}

\title[Article Title]{Founding a mathematical diffusion model in linguistics. The case study of German syntactic features in the North-Eastern Italian dialects.}


\author*[1]{\fnm{Ignazio} \sur{Lazzizzera}}\email{ignazio.lazzizzera@unitn.it}

\affil*[1]{\orgdiv{Department of Physics}, \orgname{ University of Trento}, \orgaddress{\street{via Sommarive, 14}, \city{Povo}, \postcode{38123}, \state{TN}, \country{Italy}}}


\abstract{
  The initial motivation for this work was the linguistic case of the spread of Germanic syntactic features into Romance dialects of the North-Eastern Italy, occurred after the immigration of German people in the Tyrol during the High Middle Age.
  To obtain a representation of the data over the territory suitable for a mathematical formulation, an interactive map is produced as a first step, using tools of what is called Geographic Data Science. A smooth two-dimensional surface $\mathcal{G}$ is introduced, expressing locally which fraction of territory uses a given German language feature: it is obtained by a piece-wise cubic curvature-minimizing interpolant of the discrete function that says if at any surveyed locality that feature is used or not.\newline
  This surface $\mathcal{G}$ is thought of as the value at the present time of a function describing a diffusion-convection phenomenon in two dimensions (here said \emph{tidal} mode), which is subjected in a very natural way to the same equation used in physics, introducing  a contextual diffusivity concept: it is shown that with two different assumptions about diffusivity, solutions of this equation, evaluated at the present time, fit well with the data interpolated by $\mathcal{G}$, thus providing two convincing different pictures of diffusion-convection in the case under study, albeit simplifications and approximations.\newline
  Very importantly, it is shown that the linguistic diffusion model known to linguists as Schmidt 'waves' can be counted among the solutions of the diffusion equation:  to look also at more general then the present study case, superimposing Schmidt 'waves' generated at different due times and localities and with a 'tidal linguistic flooding' just around the main region of linguistic diffusion can reproduce complexities of real events, thus probing diffusivity assumptions based on historical, local cultural, social and geographical grounds. The present work is motivating a long term research plan, seeking answers to fundamental questions of linguistics as a science, which are recalled in the article.}




\maketitle

\section{Introduction}

Linguistics is the scientific study of language, taking into account its cognitive, the social, the cultural, the psychological, the environmental, the literary, the grammatical, the paleographical, and the structural aspects \cite{Halliday}\,. Since the 20th century, it became primarily descriptive, possibly through measurable variables describing and explaining language features\cite{Martinet}\,.
\newline
A major theme nowadays is the study of the dynamics according to which variants spread in situations of multilingualism, given the greater global mobility, English as an almost universal language, the loss/disappearance of minority or local languages and dialects. From the point of view of linguistic theory, seeing how different systems influence each other is one of the biggest challenges. One wonders what phenomena are structurally impossible to pass from one language to another; by what mechanisms does a language bend its structure to make room for a variant; why do structurally different languages, that are however territorially close, influence each other (see for instance \cite{Hickey}).\newline
Diffusion of linguistic features is the subject of intense studies. For instance, to cite an interesting resulting theory, Labov in \cite{Labov 2} addresses "the problem of diffusion" underlining that linguistic variants can enter the system of adult speakers, but it is above all when it enters the system of children that they gain the possibility of spreading: adults tend not to learn from adults, it is children who learn from adults and re-functionalize sporadic variants heard by adults in their system. Another example among many to cite is the work “Minority languages in language contact situations: three case studies on language change.” by Padovan et al.\cite{Padoan}\,. \newline
Many are the factors that we expect to influence the speed and extent of diffusion when a new linguistic feature emerges within a community: first of all there is the intensity of social exchanges, while it is likely that centers (cities) and activities with greater and more prestigious social exchanges, as well as individuals with greater and more prestigious social exchanges, tend to exert a greater push in the diffusion of what J.~Schmidt called \emph{linguistic innovations} \cite{Schmidt}\,. Imagining these factors and reasoning about them, however, does not exhaust the scientific way forward: measurable variables must be devised and data analyzed with them. It is precisely to improve such way that this paper seeks to make a contribution, borrowing the theory of diffusion from physics and pointing to the geographical spread of linguistic innovations as for the main observable, with the \emph{diffusivity function} having a key parameter role, possibly the product of the rate of inter-linguistic contacts, the area of a specific locality and {\it receptance}, or the attitude to linguistic innovation, connected to environmental, social, political and even ideological factors.     
\newline
A particularly challenging target is to understand how and why certain linguistic features spread while others do not. It has triggered a large amount of works, like in \cite{Britain,HJ.Schmid,Nerbonne3}\,, just to cite few representative of them. There isn't any conclusive theory jet, apart from recognizing once more the need to introduce new, efficient, quantitative measuring variables of historical, sociological and geographical factors, in the contest of models or theories that can be proven or falsified on the basis of objective data.
\newline
Indeed, the present work is intended to provide a model that adapts to the needs of a linguistic science the so successful theory of diffusion in physics by using the methods of the recent \emph{Geographic Data Science}, in the belief that insights and techniques from such combination will be of value in improving descriptions of geographical variation in languages, and that these improvements will in turn lead to adequate answers to fundamental questions like the one just mentioned.
\newline
A particular sector of linguistics is \emph{dialectology}. A dialect is a variety of language characteristic of a particular community of individuals bound to each other by social identity; it owns its grammatical, syntactical and phonological rules, linguistic features, and stylistic aspects, however not strictly encoded and thus particularly free to evolve, in particular through the \emph{diffusion of linguistic innovations}.
\newline
As already mentioned, the trend of contemporary linguistics is to be primarily descriptive and even quantitative, like with that branch of dialectology that is called \emph{dialectometry}\,, founded by H.~Goebl years ago and aimed at finding the dynamics of a linguistic variant spread by measuring its occurrence in a given territory \cite{Goebl}\,. Indeed, as remarked for instance in \cite{Nerbonne1}, now traditionally dialectology focuses on geography, visualizing on a map the distribution of some single linguistic feature or a small number of them as a preliminary for further analysis \cite[Ch.2]{Chambers}\,.
More recently computational algorithms have been used to analyze large word data sets in different languages \cite{Goebl,Nerbonne2,Kretzschmar}\,, often taking into account various linguistic factors, such as phonetics and word frequency. A specific aim is to  measure \emph{lexical distance} \cite{Seguy}\, among languages/dialects in order to quantify their linguistic similarity or dissimilarity, as well as to detect
correlations between variations and historical-social facts \cite{Labov,Trudgill}\,.
\newline
Concerning dialectometry specifically, the present work provides new content to dialect geography, just using techniques of the Geography Data Science which, besides to offer powerful interactive visualization tools, has suggested {\it operational} observables which on the one hand have made it possible to adapt to the linguistics the diffusion theory of physics, on the other are suitable for data driven analysis.
\newline
It is taken the case of the diffusion of some German syntax features in dialects of Romance origin spoken in North-Eastern Italy, a long-term time effect of the medieval German immigration in the Alps \cite{Bidese, Tomaselli, Pescarini}. The proposed mathematical model of geographical diffusion of linguistic features promises to tell us precisely which and how phenomena have been developed in the dialects of North-Eastern Italy for the contact with German, which led to the current geographical configuration of German-like features; therefore it will offers a historical vision of the structure of the dialects in North-Eastern Italy, as well as of the exchange relationships between linguistic groups; it also will tell us according to which geographical paths, mainly, diffusion took place. \newline
The present work does not yet give answers to the fundamental linguistic questions formulated above, but intends to be the starting point of a research program to be realized in the near future, aimed at giving substance to all the expectations of understanding through a novel approach: this is its scope indeed. The program includes a large and demanding data collection campaign; it is interdisciplinary, comprising linguistics, history, sociology, data analysis and even mathematical physics; it will be conducted together with Ermenegildo Bidese (UniTn), an expert in the study of German-speaking minorities in Italy and Roman-Germanic contact in the Alps \cite{EBidese}\,. Hopefully methods and insights will be extended over different cases.

\section{Work line, main results and perspectives}

For the purpose of this work there is no need to get into any linguistic details of the sample syntactic features to be analyzed, because it is only their geographical distribution that is of interest at the moment;  they will be just referred to by labels, namely {\it meteoverbs}, {\it free\_Subj\_hum}, {\it free\_Subj\_not\_hum}, {\it Wh\_encl}. For the local dialect of each of the surveyed locality in the North-Eastern Italy it is stated if a German feature is detected or not, so that a discrete function of the geographic coordinate of the localities is set to the values “1” or “0” (“yes” or “not”): such function is called $\textbf{\textit{g\_index}}_l$, with the subscript $l$ to denote a locality. 
.\newline
The first step will be the introduction of a {\it smooth} two-dimensional surface $\textbf{\textit{g\_index}(\textit{longitude, latitude})} \equiv \textbf{\textit{g\_index}(\textit{x, y})}$, that gives which fraction of a small area around the geographic point (x,y) uses some specific German language feature: in particular, given a surveyed locality $l$ of area $A_l$, it equals the discrete function $\textit{g\_index}_{\,l}$ over the whole $A_l$; it is obtained by a piece-wise cubic curvature-minimizing interpolant of the discrete function itself and will be a continuous and continuous derivative function almost everywhere over the North-Eastern Italy region. 
\newline
Then, {\it contour lines} and corresponding {\it gradient lines} on the  $\textit{g\_index}$ surface will be introduced, indeed a {\it gradient field}, the latter to be used to define a {\it linguistic features flux}, the key passage to dynamical models of {\it linguistic features diffusion}. Very importantly, linguistic feature fluxes are defined and correspondingly measurable following specific trajectories, allowing valuable terms of comparison among different features and along different paths.
\begin{figure}[!ht] 
  \centering
  \includegraphics[width=\textwidth]{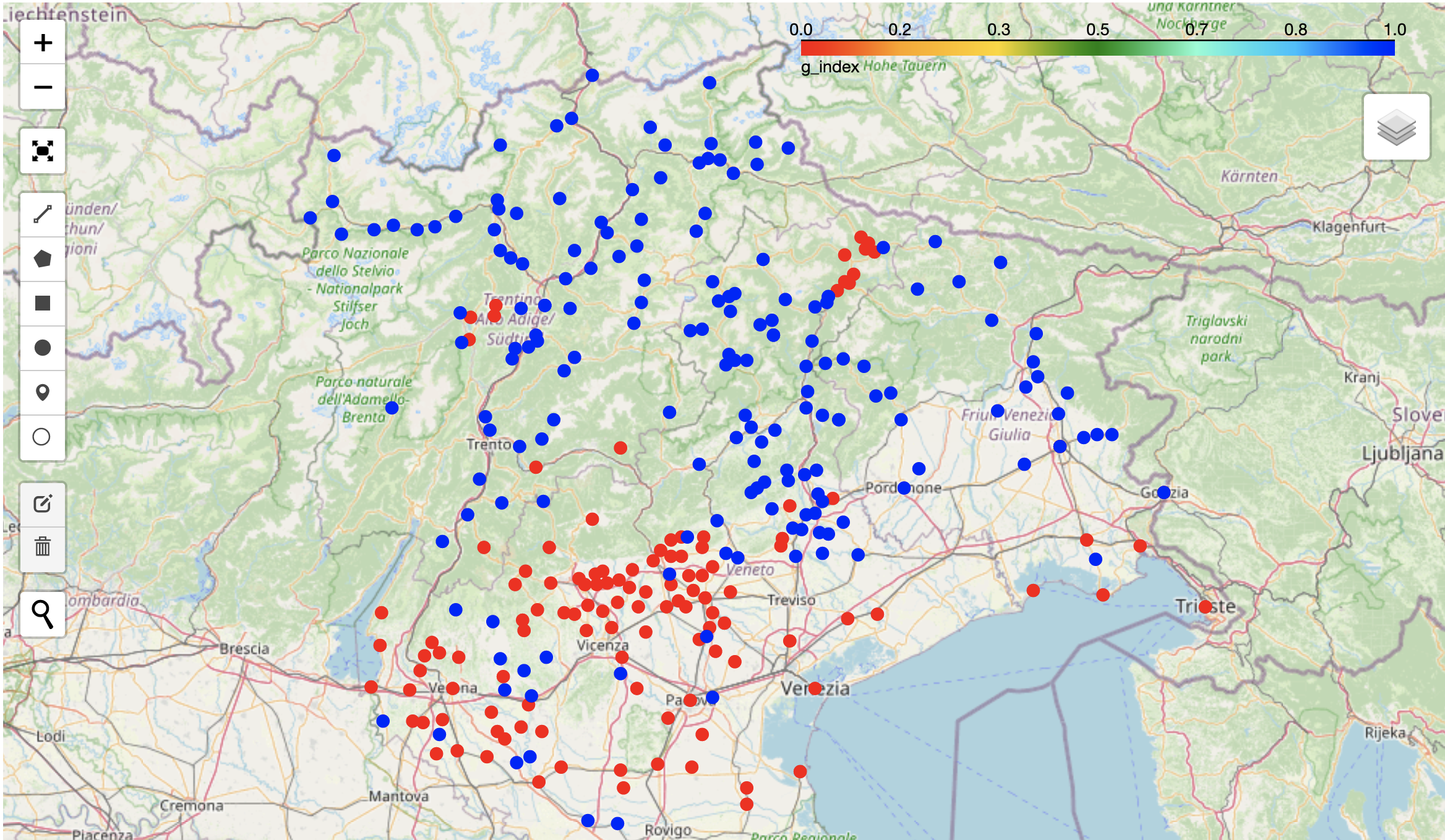}
  \caption{The $\textit{g\_index}$ raw data representation,  blue corresponding to $\textbf{\textit{g\_index}}_l = 1$ (the German feature is found in the dialect at $l$) and red to $\textbf{\textit{g\_index}}_l = 0$ (the German feature is not found in the dialect at $l$).}
  \label{raw}
\end{figure}
As a scientific visualization tool, the package ``folium Python'' \cite{folium,geographicdata} is adopted and ``Jupyter Notebook'' is used as the software development environment \cite{jupyter}.
\newline
Fig.~\ref{raw} shows the raw data for {\it meteoverbs}, with the {\it blue} to indicate $\textbf{\textit{g\_index}}_l = 1$ (the German feature is found in the dialect at $l$) and the {\it red} to indicate $\textbf{\textit{g\_index}}_l = 0$ (the German feature is not found in the dialect at $l$).
\newline
Often in the sequel the details of the procedure are presented with reference to {\it meteoverbs} in particular.
\newline
The linguistic data were georeferentially extracted and reprocessed by Romano Madaro (University of Trento) from the VinKo (Varieties in Contact) platform. VinKo is a spoken corpus based on crowd-sourced audio recordings that has been designed by the Universities of Trento and Verona to provide relevant linguistic information about the minority languages and dialects spoken in the area between Innsbruck and the Po Valley (www.vinko.it) \cite{Rabanus et al}  and financed by the European Seventh Framework Program for Research, Technological Development and Demonstration - Project code: 613465. Project name: AThEME (Advancing the European Multilingual Experience).
\newline
As the main result of the work, mathematical formulations for linguistic diffusion-convection and Schmidt 'waves' models are provided. It is shown that a basic solution of the diffusion-convection equation, the \emph{complementary error function} (\emph{erfc}), as well as a piece-wise linear one, well fit the data given in a devised, original elaboration. Most importantly in the author's opinion, it is shown that the beautiful J.~Schmidt's image of the \emph{language innovations} spreading as circular \emph{waves} is substantiated through specific solutions of the diffusion equation, of which an analytic example is given explicitly: in a paper in preparation a whole set of solutions is provided, dependent on diffusivity laws ranging as the inverse of any power of the distance from a centre of diffusion; analysis of the data through fits to the data will allow, on the one hand, to approach the question of language contact from a completely different perspective than that currently prevailing in linguistics and, on the other hand, to model the historical colonisation of the higher Alpine regions by German immigrants in the Middle Ages, in a way that is substantially different from and complementary to those used in history and sociology.

\section{Definitions, procedure and a natural model of linguistic features diffusion}
As mentioned above, a piece-wise cubic curvature-minimized interpolant is used to define the two-dimensional continuous and continuous derivative function $\textit{g\_index}$\,, such that for a surveyed locality $l$ of area $A_l$ one gets
\begin{equation*}
  \int_{A_l} \text{d}x\,\text{d}y\;\,\textit{g\_index}\,(x,y) =
  A_l\;{\textit{g\_index}}_l
\end{equation*}
The interpolant is obtained triangulating over the data by an algorithm that makes use
of  a cubic curvature minimizing Bezier polynomial on each triangle in a Clough-Tocher scheme \cite{Alfeld,Farin}: the Python {\it scipy}  {\it CloughTocher2DInterpolator} module \cite{scipy_interpolate} was used.
\newline
The $\textit{g\_index}$ surface could be given a simple interpretation: at a given locality $l$ of infinitesimal area $a_l$ it would give the fraction of territory whose local dialect adopts the given Germanic syntactic feature, so that, shortening the notation $\textbf{\textit{g\_index}(\textit{x, y})}$ to $\mathfrak{G}(x,y)$ in formulas from now on, the integral
\begin{equation*}
  \int_\text{North-Eastern Italy} \de{x}\, \de{y}\; \mathfrak{G}\,(x, y)
\end{equation*}
would give the area of the North-Eastern Italy where that one Germanic syntactic feature in used. 
\newline
There is a problem, however, coming from the requirement that the interpolant be continuous and continuous differentiable everywhere: points on areas contained in closed contour lines of level 1 get $\textit{g\_index}$ higher then 1 and points on areas contained in closed contour lines of level 0 get negative $\textit{g\_index}$, a circumstance indeed inconsistent with the above proposed interpretation. The solution is to impose that the $\textit{g\_index}$ surface with continuity becomes flat and equal 1 on those areas contained in closed contour lines of level 1 and flat and equal to 0 on those areas contained in closed contour lines of level 0; of course the derivative will not be continuous everywhere. 

\subsection{ The contour lines}

\begin{figure}[!ht]
  \centering
  \includegraphics[width=0.9\textwidth]{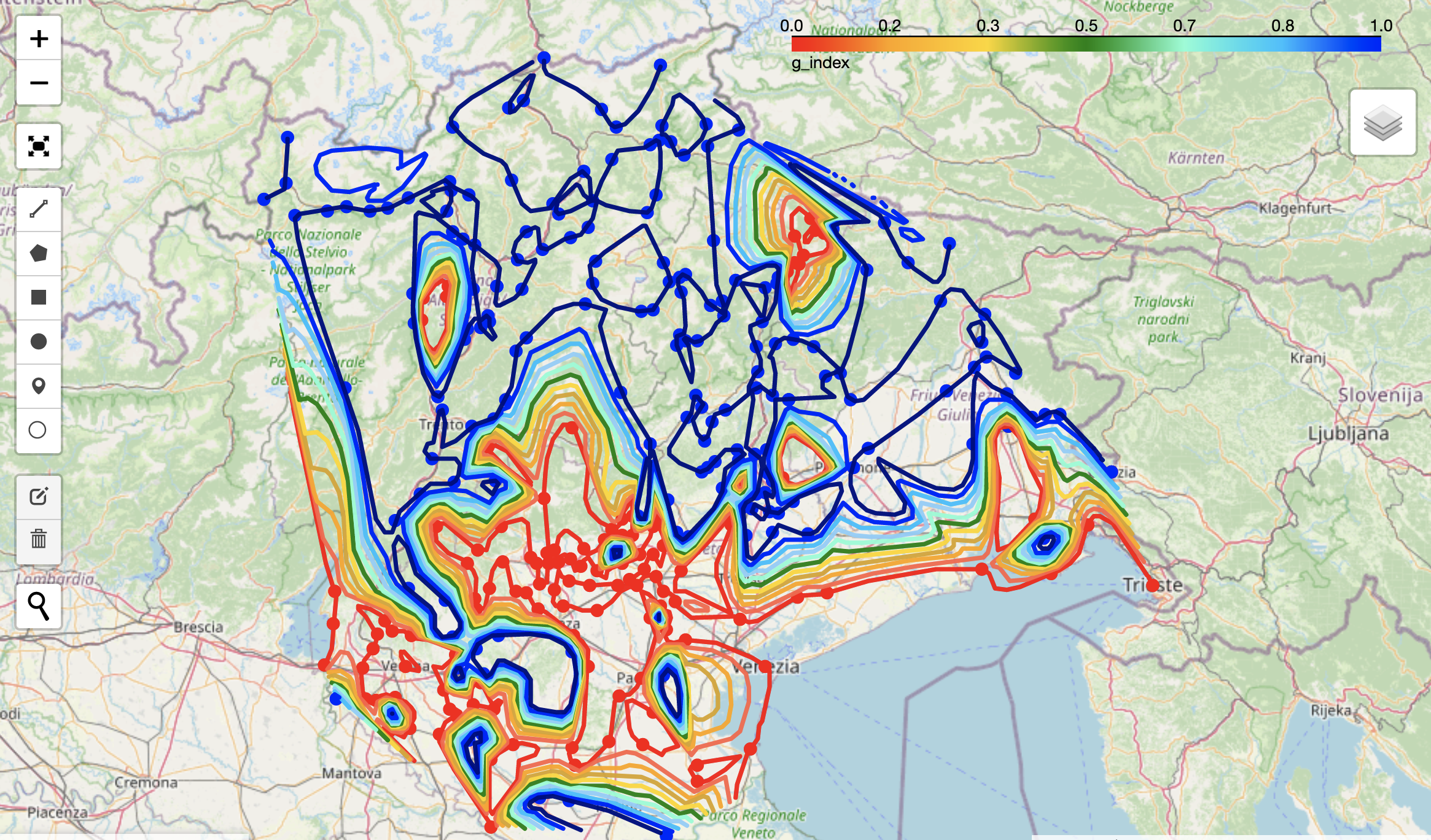}
  \caption{The $\textit{g\_index}$ contour lines: {\it meteoverbs}}
  \label{contours}
\end{figure}

\begin{figure}[!ht]
  \centering
  \includegraphics[width=0.9\textwidth]{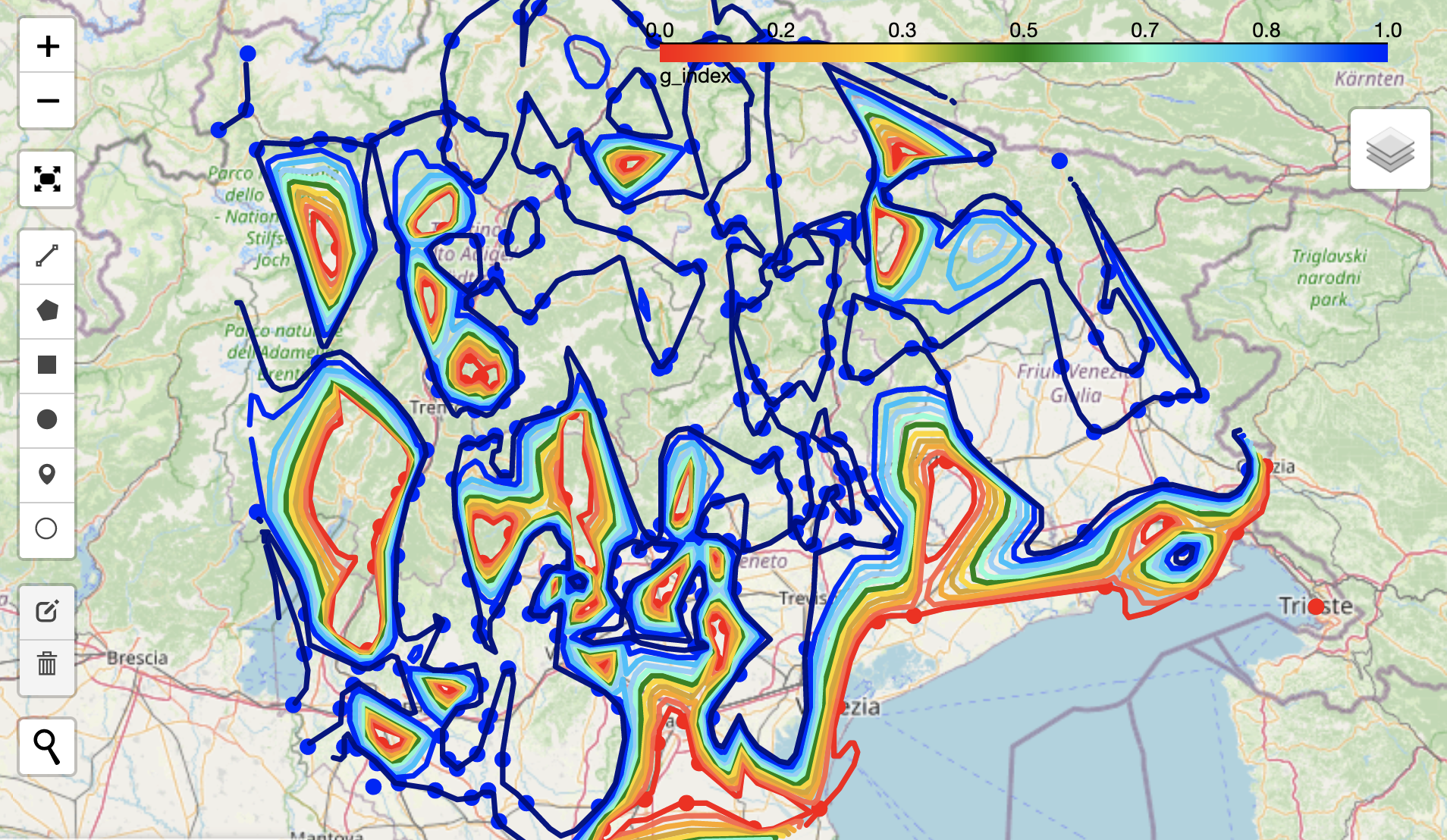}
  \caption{The $\textit{g\_index}$ contour lines: {\it free\_Subj\_hum}}
  \label{free_Subj_hum}
\end{figure}

\begin{figure}
  \centering
  \includegraphics[width=0.9\linewidth]{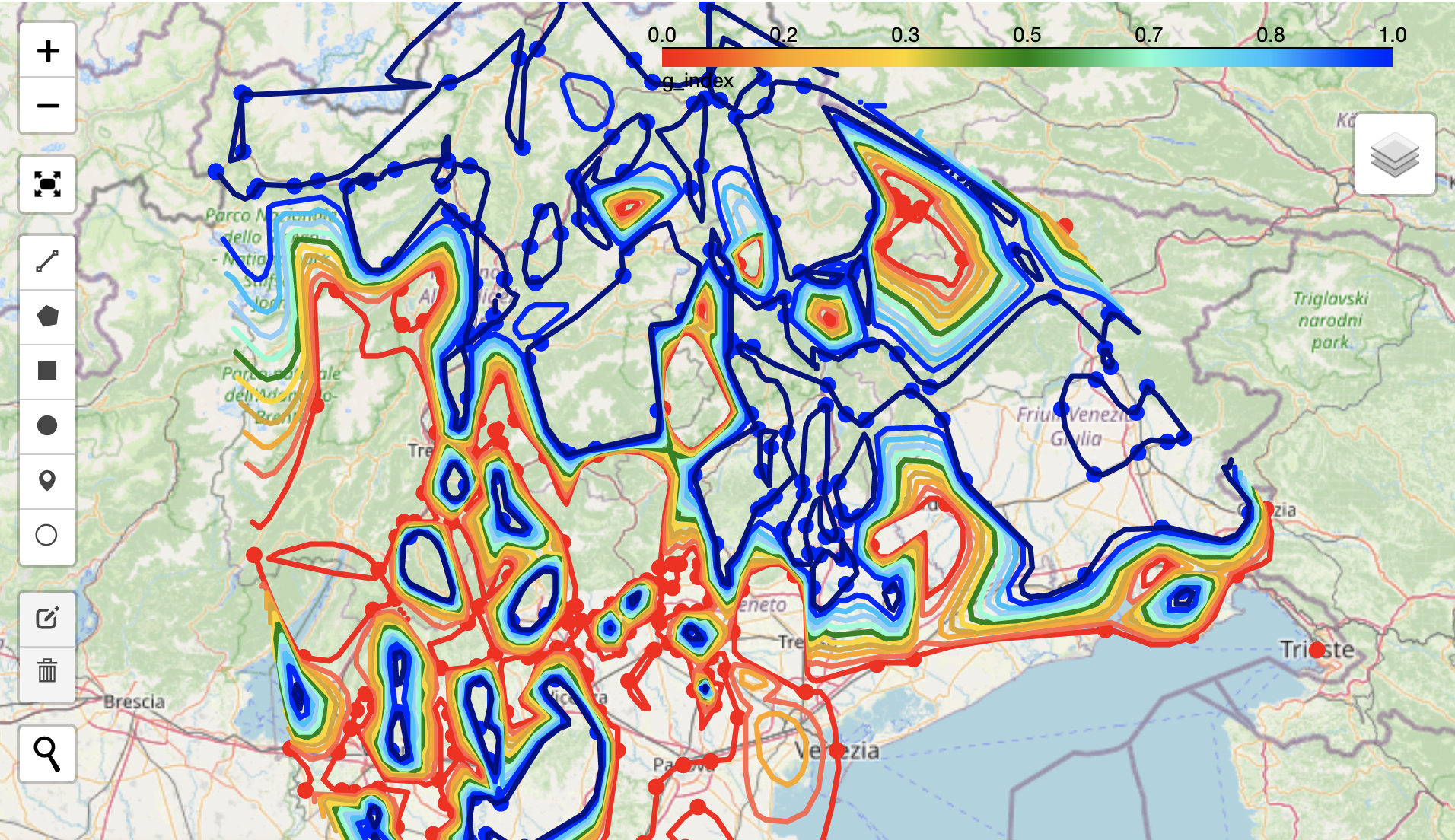}
  \caption{The $\textit{g\_index}$ contour lines: {\it free\_Subj\_not\_hum}}
  \label{free_Subj_not_hum}
\end{figure}

\begin{figure}[!ht]
  \centering
  \includegraphics[width=0.9\textwidth]{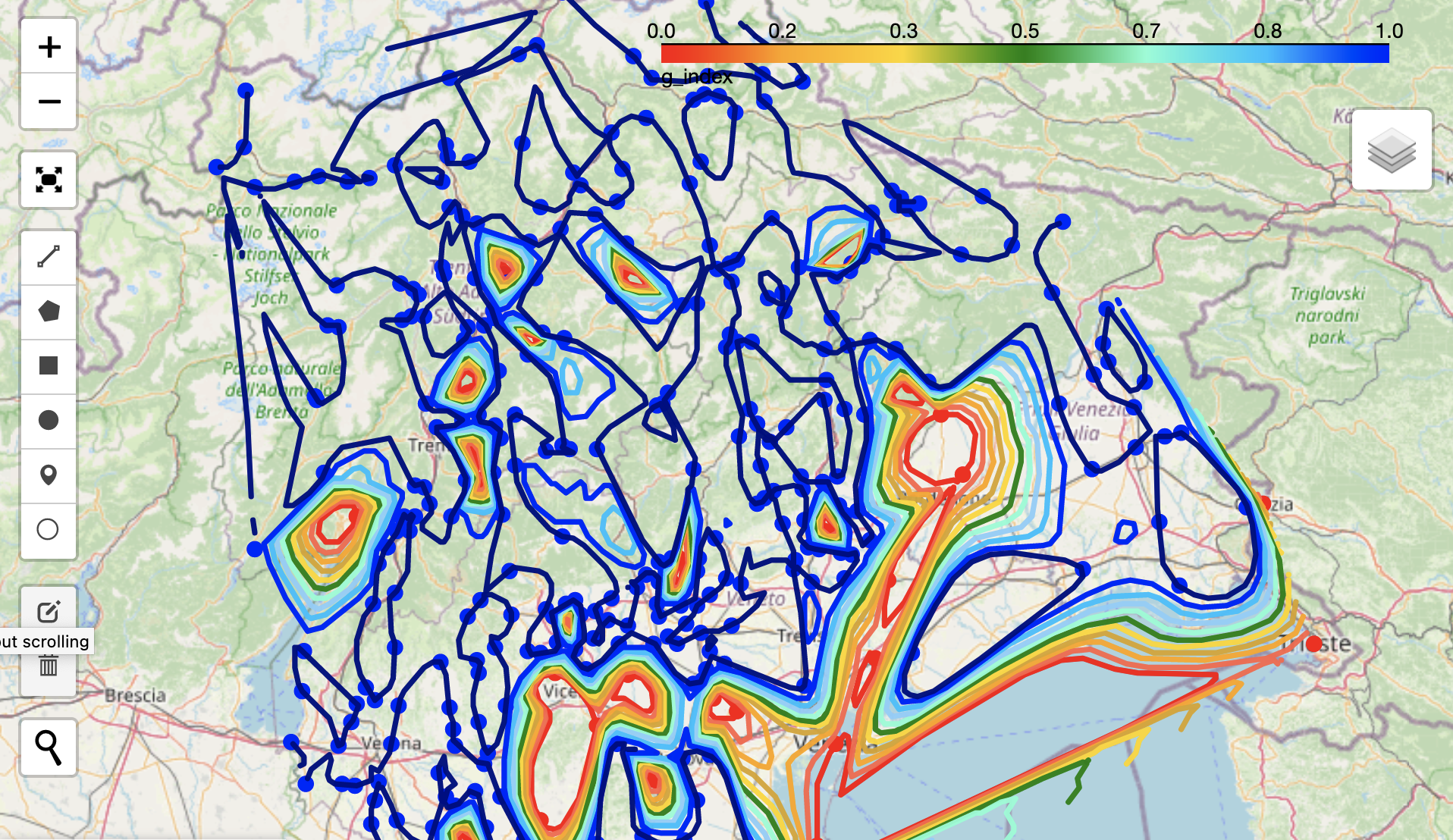}
  \caption{The $\textit{g\_index}$ contour lines: {\it Wh\_encl}}
  \label{Wh_encl}
\end{figure}

The contour lines are easily obtained on the $\textit{g\_index}$ surface using a specific module of the Python Matplotlib Pyplot package \cite{pyplot}.\; In \Cref{contours,free_Subj_hum,free_Subj_not_hum,Wh_encl} are shown the contour lines of levels from 1.0 to 0.0 in sub-intervals of 0.1 for the four features of the case study.

\subsection{Gradient field of the g\_index and the equation of diffusion}

To remind linguists not familiar with the notion of gradient of a scalar field, like the $\textit{g\_index}$ function is, take a point $p(g)$ on a contour line of level $g$ and consider an adjacent contour line of level $g+\delta g$ with $\delta g \ll\, 1$; identify the shortest path from the point $p(g)$ to the contour line $g+\delta g$ and let $\delta s$ be the length of that path: in the limit for $\delta s\, \rightarrow \,0$ (and thus $\delta g\, \rightarrow \,0$) the ratio $\delta g/\delta s$  is the gradient of  $\textit{g\_index}$ at point $p(g)$\,; one easily realizes that the shortest path from $p(g)$ to the contour line $g+\delta g$ is orthogonal to the latter. In summary, the gradient of $\textit{g\_index}$ is a vector that says in which (geographic) direction and how much $\textit{g\_index}$ varies maximally; it is denoted by 
$$ \vec{\nabla} \mathfrak{G}\,(x,y)\,,$$
where $x,\,y$ are the coordinates of $p(g)$\,.\\ 
Taking an infinitesimal displacement $\de{s}$ along the direction of the gradient, recalling that $\lvert\vec{v}\rvert$ means magnitude of the vector $\vec{v}$, then
$$\lvert\vec{\nabla}\mathfrak{G}\,(x,y)\rvert \; \de{s}$$
tells how much the territorial fraction, where some Germanic syntactic structure is adopted, varies over that given displacement: hereinafter, with reference to each Germanic linguistic feature, the portion of territories where that feature is adopted will be referred to as {\it g\_like}; its complement will be said {\it i\_like}.
\newline
Getting back to \cref{contours,free_Subj_hum,free_Subj_not_hum,Wh_encl}\,, one sees a major strip of territory where the gradient field is not null, marked by (almost equidistant) lines of different colours (that is of different g\_index), which develop serpentine roughly along the geographical parallels: it will be referred to as a \textbf{\emph{gradient front of linguistic diffusion}}. Most of the areas north of such {\it gradient front} and south of Tyrol saw the rooting of Germanic linguistic features in the local Romance dialects: it would be interesting to find an explanation for the apparent differences in the expansion areas relatively to the different syntactic constructs. Anyway, quite apart of that, it seems that for all the four different features here considered, the contour intervals along each {\it gradient front} have almost same transverse geographic width and are roughly constant, as will be shown quantitatively in the sequel.
\newline
\newline
It is likely that the rate at which an {\it i\_like} territory changed into {\it g\_like} was determined by the rate $\nu$ of inter-linguistic contacts, the area of a specific locality and a related {\it receptance} $\rho$ connected to environmental, social, political and even ideological factors. Following various diffusion theories in physics, one would introduce a function $\eta(x,y; \mathfrak{G})$, to be called {\it diffusivity}, that could be the product of the parameters just mentioned, thus having dimension of a length squared per time unit; then one would introduce a {\it local directional flux} of linguistic feature as
\begin{equation}
  \label{Fick1}
  \vec{\Phi}  \;=\; -\,\eta\,\vec{\nabla} \mathfrak{G} \,.
\end{equation}
It is worth to say that Eq.~(\ref{Fick1}) just re-frames the well known Fick's first law of diffusion in physics.\newline
To get an intuitive idea of what one is going to do, think of an imaginary colouring fluid spreading from an extended source over a forward territory: areas of that territory that were originally red because genuinely \emph{i\_like}, are given the colour blue when they acquire Germanic syntactic structures, as a novel habit caused by frequent contacts with nearby German-speaking people and by attitude to linguistic innovation.\\
More precisely, an area would be changed in colour by the formula \\
\centerline{\textit{\textbf{
      g\_index\;$\bullet$\,blue + (1\,-\,g\_index)\;$\bullet$\,red}}\,,}\\
according to which fraction of that area changed to \emph{g\_like} from \emph{i\_like} and which fraction remained \emph{i\_like} (this formula would be the colour code of the contour lines of \cref{contours,free_Subj_hum,free_Subj_not_hum,Wh_encl}\,; $\vec{\Phi}$ would come to be interpreted, symbolically, as the local directional flux of the colouring fluid).
\newline
From the Gauss theorem in two dimensions, calling $A$ a closed geographic area and $\Gamma=\text{Fr}(A)$ its border (or {\it frontier}), one has
\begin{equation}
  \int_A \de{x}\,\de{y}\; \vec{\nabla}\cdot\vec{\Phi} \;=\;
  \oint_\Gamma \de{\vec{l}}\cdot\vec{\Phi} \;=\;
  -\,\frac{d}{dt} \int_A \de{x}\,\de{y}\,\mathfrak{G} \;=\;
  -\,\int_A \de{x}\,\de{y}\, \pdv{\mathfrak{G}}{t} \;,
  \label{conservazione del fluido colorante in forma integrale}
\end{equation}
where in the metaphor of the colouring fluid the second integral would represent, by definition, the total flux of the outgoing fluid across the border of $A$, equal to the rate of decrease of the total of the $\textit{g\_index}$ over $A$ as expressed in the third term of the chain of equality's: since this term has dimension of area per time unit, from the first term of (\ref{conservazione del fluido colorante in forma integrale}) it follows that the {\it local directional flux} $\vec{\Phi}$ has dimension of length per time unit; consequently, from Eq.~(\ref{Fick1}) it can be seen that $\eta$ has dimension of a length squared per time unit as expected. Following up on the image of the colouring fluid, $\vec{\Phi}$ would represent {\it how far the colouring fluid in the time unit advances, orthogonally crossing a line segment of unit length.}
\newline
Since the area $A$ is arbitrary, from the first and the last terms of (\ref{conservazione del fluido colorante in forma integrale})\, with usual arguments in mathematics one obtains the following {\it continuity equation}
\begin{equation}
  \pdv{\mathfrak{G}}{t} = - \vec{\nabla} \cdot \vec{\Phi} \,,
  \label{continuity equation}
\end{equation}
which gives in differential form the conservation law stated in
(\ref{conservazione del fluido colorante in forma integrale})\,. 
\newline
Using in Eq.~(\ref{continuity equation})\, the definition of $\vec{\Phi}$ given in Eq.~(\ref{Fick1})\,, the second Fick's law of diffusion follows, i.e.:
\begin{equation}
  \label{diffu eq}
  \pdv{\mathfrak{G}}{t} =
  \vec{\nabla} \cdot \left(\eta\, \vec{\nabla} \mathfrak{G} \right) \,.
\end{equation}

\subsection{ Gradient lines and approximate projection onto one-dimension}
A {\it gradient line} is a path built by successive infinitesimal steps along the local
\begin{figure}[!ht] 
  \centering
  \includegraphics[width=0.9\textwidth]{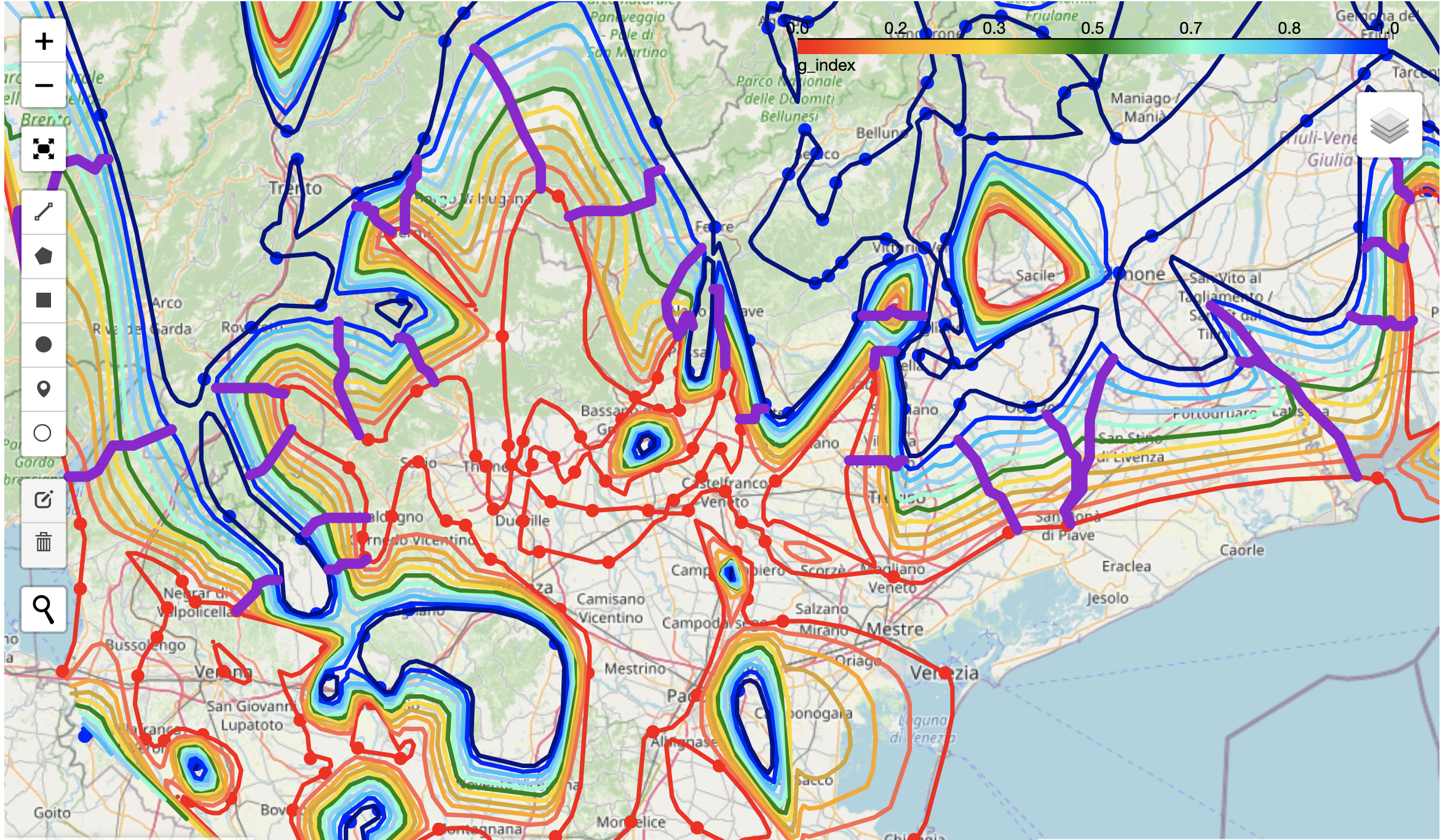}
  \caption{Some force lines (in violet) from 0.9 to 0.0}
  \label{force_lines}
\end{figure}
direction of the gradient field, tangentially to the gradient vector; it traverses orthogonally each contour line; indeed it is a {\it steepest descent} curve. A custom code automatically builds a number of sample gradient lines, shown in violet in Fig.~\ref{force_lines}\,: for practical reasons they go from initial points chosen (almost) randomly on the contour line of level 0.9 and reach the contour line of level 0.0\,; really, they are approximately gradient lines, in the sense that they follow the steepest descent in straight segments of 0.1 contour intervals.
\newline
In general numerical methods must be used to solve Eq.~(\ref{diffu eq})\,; however, in the following, an approximate projection onto one-dimension, often solved analytically, is used. 
\newline
By its very definition, the gradient of a function is in fact the directional derivative along a gradient line. Denoting by $\vec{\gamma}:\Rbb \rightarrow \cal{M}$ a gradient line over the (geographic) map $\cal{M}$ and by
$\vec{u}(s)$ the tangent versor to $\vec{\gamma}$ in $s$, assuming $\vec{\gamma}$ to be almost a straight line, that is $\vec{u}$ to be almost independent of $s$\,, it is proved in \ref{diffu to one dim} that Eq.~(\ref{diffu eq}) can be written in the form
\begin{equation}
  \label{one D diffu eq}
  \pdv{\mathcal{G}_\gamma}{t} \,=\,
  \pdv[order={1}]{}{s}\left(\eta_\gamma\,
  \pdv[order={1}]{\mathcal{G}_\gamma}{s}\right)\,.
\end{equation}
An important investigative tool is provided by a custom code, that builds gradient lines choosing manually the starting point: with just {\it clicking} on the interactive map at a point of interest, the corresponding gradient line is automatically generated, together with information about its geometry. This is quite useful to investigate, say, the German language features in Italian dialects along some specific geographic {\it corridor} that can be of interest, like along the Adige Valley or similar, for the case study here.
\newline
In the sequel the average $\mathcal{G}$ of $\mathcal{G}_\gamma$\,'s over a number of gradient paths $\gamma$ will be taken: exactly in this sense the subscript $\gamma$ will be dropped.

\subsection{ Looking for suitable solutions of the one-dimensional diffusion equation}
\begin{figure}[!ht] 
  \centering
  \includegraphics[width=0.6\textwidth]{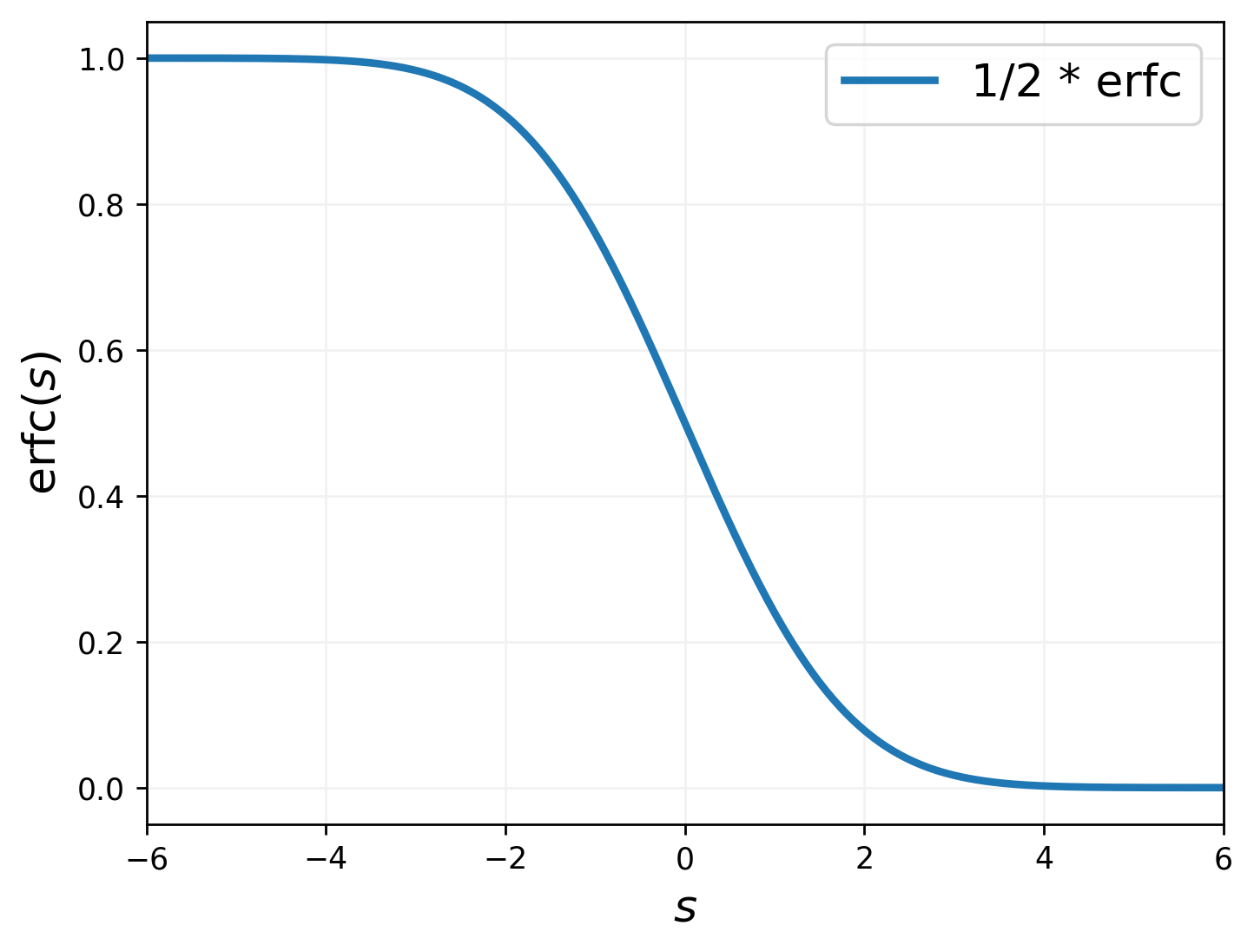}
  \caption{a sample of the complementary error function (erfc)}
  \label{erfc.png}
\end{figure}

As already seen, the diffusivity $\eta$ has dimension of length square per time unit, so that for instance, quite apart from being a constant, it could have the simple form 
\begin{equation}
  \label{Bolz linear}
  \eta = \frac{s^2}{4\,t}\,,
\end{equation}
which in fact happens to generate a solution piece-wise linear in $z$, that will be used in the sequel: seen as the radial diffusivity from a point source on a plane, this behavior  of $\eta$, quadratic in the radius $s$, should not be surprising, because the ratio with the surface of the circle of that radius just behaves as a constant divided by the time $t$\,.\\
It was first suggested by L.~Boltzmann in 1894 \cite{Boltzmann} that for all similar functional laws of $\eta$ or when $\eta$ depends on $\mathcal{G}$ only, the transformation 
\begin{equation}
  \label{Bolz subst}
  s,\,t\; \rightarrow \; z = \frac{s}{2\sqrt{t}}\,,
\end{equation}
by the chain rule of derivatives
\begin{equation}
  \label{B subst}
  \pdv{\mathcal{G}}{t} = - \frac{s}{4\,t\,\sqrt{t}}\,\odv{\mathcal{G}}{z} \,,
  \qquad\quad
  \pdv{}{s}\left(\eta\,\pdv{\mathcal{G}}{s}\right) = \frac{1}{4\,t}\,
  \odv{}{z}\left( \eta\,\odv{\mathcal{G}}{z} \right)\,,
\end{equation}
reduces Eq.~(\ref{one D diffu eq}) to the ordinary (in general non-linear) differential equation
\begin{equation}
  \label{ord diff eq}
  -\,2\,z\,\odv{\mathcal{G}}{z} \,=\,
  \odv{}{z}\left( \eta\left[z;\,\mathcal{G}\right]\,\odv{\mathcal{G}}{z} \right)\,.
\end{equation}
\newline
Assuming $\eta$ is a constant then, a solutions of this equation is given in \ref{error funct} and will be used later: it is proportional to the so called {\it error function complement} (erfc) and has the following analytic expression
\begin{equation}
  \label{erfc}
  \mathcal{G}(s,\,t) \,=\, \frac{1}{2}\, 
  \left(1 \,-\,\frac{2}{\sqrt{\pi}}\,\int_0^{\;s/(2\,\sqrt{\eta\,t})}\de{u}\;e^{-u^2}\right)
  \,\equiv\, \frac{1}{2}\,\erfc{\left( \frac{s}{2\,\sqrt{\eta\,t}}\right)} \,.
\end{equation}
For $2\,\sqrt{\eta\,t}\,=\,1$\,) it is plotted in Fig.~\ref{erfc.png}\;; it exhibits an inflection point when its argument equals zero; the tangent in this point approximately overlaps the curve for a good stretch around.
\subsection{A {\bf \emph{convection}} component in the diffusion of linguistic innovations}
It is likely that the diffusion of German linguistic features was not only due to exchanges of {\it i\_like} dialects speaking people with the nearby population of the Tyrol, but also to internal mobility, connecting people who already used {\it g\_like} structures with people who still didn't, thus creating an additional transport dynamics of linguistic innovation.  This dynamics, similar to what is called in physics {\bf \emph{convection}},  is introduced in Eq.~(\ref{diffu eq}) by means of a term proportional to the gradient of the {\it g\_index} function; namely
\begin{equation}
  \label{diffu eq extended}
  \pdv{\mathfrak{G}}{t} =
  \vec{\nabla} \cdot \left(\eta\, \vec{\nabla} \mathfrak{G} \right) \,-\,
  \frac{\lambda}{\tau}\, \vec{u}\cdot\vec{\nabla}\mathfrak{G}\,,
\end{equation}
where $\vec{u}$ is a constant versor and $\lambda$ has the meaning of average distance at which the \textbf{\emph{gradient front}} of the  $\textit{g\_index}$ function had propagated from the border of Tyrol in the time interval $\tau$, roughly starting from the end of the period of the German immigration to the Alps. 
The important point is that, setting
\begin{equation*}
  \bcheck{\mathfrak{G}}(t,\,\vec{r}) \,=\,
  \mathfrak{G}\left(t,\,\vec{r} - \frac{\lambda}{\tau}\,\vec{u}\,t\right)\,,
\end{equation*}
Eq.~(\ref{diffu eq extended}) is satisfied if and only if
\begin{equation*}
  \pdv{\bcheck{\mathfrak{G}}}{t} =
  \vec{\nabla} \cdot \left(\eta\, \vec{\nabla} \bcheck{\mathfrak{G}} \right)\,,
\end{equation*}
for
\begin{equation*}
  \pdv{\bcheck{\mathfrak{G}}}{t} = \pdv{\mathfrak{G}}{t} -
  \frac{\lambda}{\tau}\,\vec{u}\cdot \vec{\nabla} \mathfrak{G}\,,\qquad
  \vec{\nabla} \cdot \left(\eta\, \vec{\nabla} \bcheck{\mathfrak{G}} \right)\,=\,
  \vec{\nabla} \cdot \left(\eta\, \vec{\nabla} \mathfrak{G} \right)\,.
\end{equation*}
It means that, with the convection term, a \emph{gradient front} would move as a whole with constant speed $\vec{v} = (\lambda/\tau)\,\vec{u}$\,.\newline
A modified 1D version is
\begin{equation}
  \label{one D diffu eq extended}
  \pdv{\mathcal{G}}{t} \,=\,\pdv{}{s}\left(\eta\,\pdv{\mathcal{G}}{s}\right)
  \,-\, \frac{\lambda}{\tau}\,\left.\odv{f(t')}{t'}\right|_{t'= t/\tau}\,
  \pdv{\mathcal{G}}{s}\,.
\end{equation}
.\newline
In this case, as shown in detail in \ref{wave solution}\,, one has that:
\begin{align*}
    &\text{\bf if}\quad {\mathbf g\left(\frac{s-s_0}{2\sqrt{t}}\right)} \quad
    \text{\bf is a solution of Eq.~(\ref{one D diffu eq})}\,,\\
    &\text{\bf then}\\
    &{\mathbf g\left(\frac{s - s_0 - \lambda\,f(t/\tau)}{2\sqrt{t}}\right)}\quad
    \text{\bf is solution of Eq.~(\ref{one D diffu eq extended})}\,,
\end{align*}
where $s_0$ is an arbitrary distance constant.\newline
So, from a solution of Eq.~(\ref{one D diffu eq}) (or equivalently
Eq.~(\ref{ord diff eq})) one gets a solution of Eq.~(\ref{one D diffu eq extended})\,, which manifestly represents a {\it gradient front} advancing and lengthening simultaneously. In \ref{wave solution} two types of convection terms are introduced, the first implying a constant progression of the gradient front, the second implying that the {\it gradient front} initially progresses at a speed approximately constant, then slows down, stops and finally slowly begins to regress; both examples are given in order to show the flexibility of the model in relation to the possibility of interpreting an historical evolution of linguistic spread, when historical data were in fact available. In the sequel both choices will be illustrated in detail.  
\newline
Everything of the \emph{diffusion-convection} model looks appropriate for the present case study: for each German linguistic feature the elaborated data is characterized by a {\it gradient front} which must have propagated over time, because one can suppose that initially it was roughly on the border of Tyrol, whereas today, although with considerable variance over the territory, it is on average around more then 50 Km away; furthermore, the transversal width of the front itself must have been initially very narrow at the Tyrolean border, whereas today it is on average around more then 12 Km wide: Eq.~(\ref{one D diffu eq}) governs this widening, while the convection term term in Eq.~(\ref{diffu eq extended}) governs the progression of the {\it gradient front} at the average velocity which can be estimated to be 50Km/1000yr or more.

\subsection{ A brief historical note}
\label{historic}
It is plausible that the bulk of Germanic immigration into the Alps ended around 1027, year of the establishment of the bishoprics of Trent and Brixen by the Holy Roman Emperor Conrad II the Salian. However, presumably since ever, but even more so in later centuries, contacts and presences of people and groups of Italic languages occurred in the South Tyrol, thanks to economic practices and the spread of the notarial system, which was mainly present in the Trentino. Commercial contacts kept relations with Venice, to which valuable wood used for shipbuilding was exported, as from the two commercial metropolises of southern Germany, Nuremberg and Augsburg. \cite{Euregio,Zieger,Obermair,Riedmann}.
\newline
On the basis of these historical considerations, it sounds plausible that the process of spread of the Germanic language structures into the dialects of North-Eastern Italy began around the year 1000, that justifies the estimation of the parameter $v$ in Eq.~(\ref{diffu eq extended}) as 50Km/1000yr or more.

\section{The data analysis}
\label{The data analysis}
The available data represents just a {\it snapshot} to nowadays of some Germanic syntactic structures present in the dialects of the North-Eastern Italy. What can be done and is actually done is to check if solutions of the diffusion-convection equation, evaluated at the present time, fit the data.\newline
In practice, the average of the $\textit{g\_index}$ over a number of gradient paths will be studied, that is the average of a number of $\mathcal{G}_\gamma$\,'s, each taken along its own $\gamma$\, path. The procedure is as follows. 
\newline
A gradient path is approximated by the sequence of nine steepest descent straight segments, starting somewhere on the 0.9 contour line, then descending through the contour lines of level $0.1\!\times\!(9-i)$, with $i = 1,\dots 9$\,.\newline
Given the k-th of $N$ different gradient paths, on it one has nine segments of length $\delta_k^{(j)}$ respectively, with $j = 1, 2, \dots, 9$\,.
\newline
Then, one sets the averages 
\begin{equation}
  \label{le delta}
  \avg{\delta^{(j)}} \defeq
  \frac{1}{N}\,\sum_{k=1}^N \delta_k^{(j)}\,, \qquad  j = 1, 2, \dots, 9\,.
\end{equation}
\newline
\begin{figure}[ht!] 
  \centering
  \includegraphics[width=0.75\textwidth]{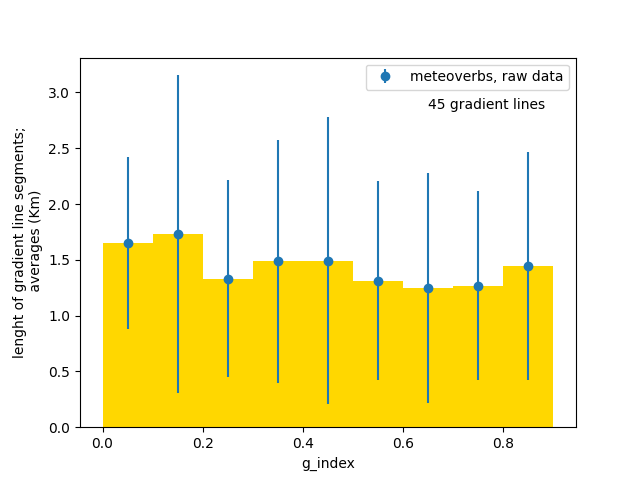}
  \caption{Average geographic width of the 0.1 wide contour intervals}
  \label{lenghts_raw_data_45.png}
\end{figure}
\newline
Fig.~\ref{lenghts_raw_data_45.png} illustrates the result for 45 gradient paths. First of all one should realize that the vertical bars should not be considered as uncertainty in the sense of statistics, but just the measured variance; quite apart from this, it is apparent that the gradient segments have all approximately the same length in the average.\newline
To probe this quantitatively, the $\textit{g\_index}$ is plotted as a function of the average distance $\avg{d^{(j)}}$ of a point at level $j$ from the contour line of level 0.9 along the (mean) gradient path. Using the definitions of $\avg{\delta^{(j)}}$ given in Eq.(\ref{le delta}) and setting $\avg{\delta^{(9)}} = 0$\,, one has
\begin{equation}
  \label{le distanze}
  \avg{d^{(j)}}\,\defeq\, \sum_{l=0}^j \avg{\delta^{(9-l)}} \qquad j=0, 1, 2, \dots 8
\end{equation} 
\newline
\begin{figure}[!ht]
  \begin{minipage}{0.49\textwidth}
    \centering
    \includegraphics[width=0.95\textwidth]
                    {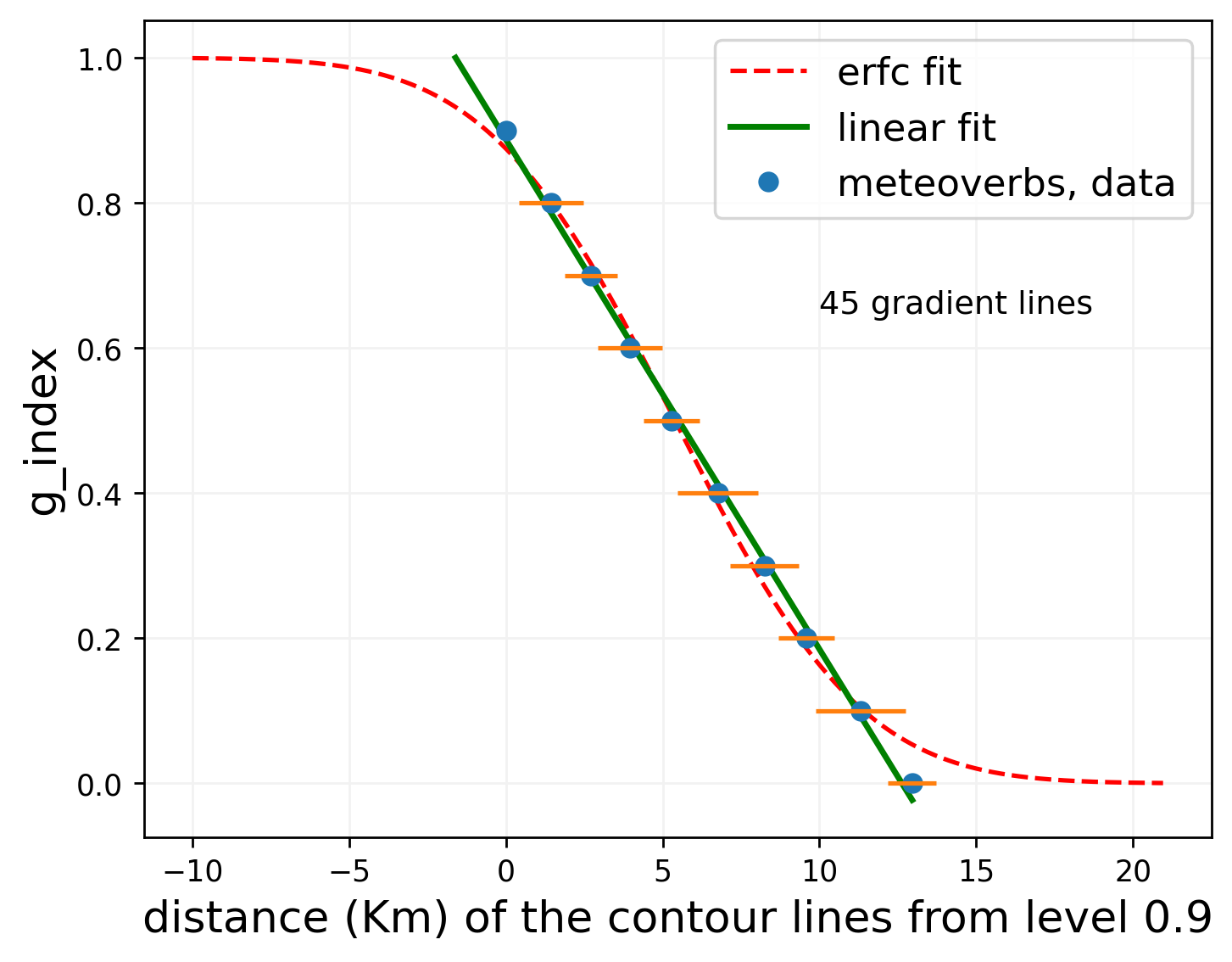}
                    \caption{Fit to {\it meteoverbs}}
                    \label{met g_index_vs_gradient_line_distances_45_Lin.png} 
  \end{minipage}\hfill
  \begin{minipage}{0.49\textwidth}
    \centering
    \includegraphics[width=0.95\textwidth]
                    {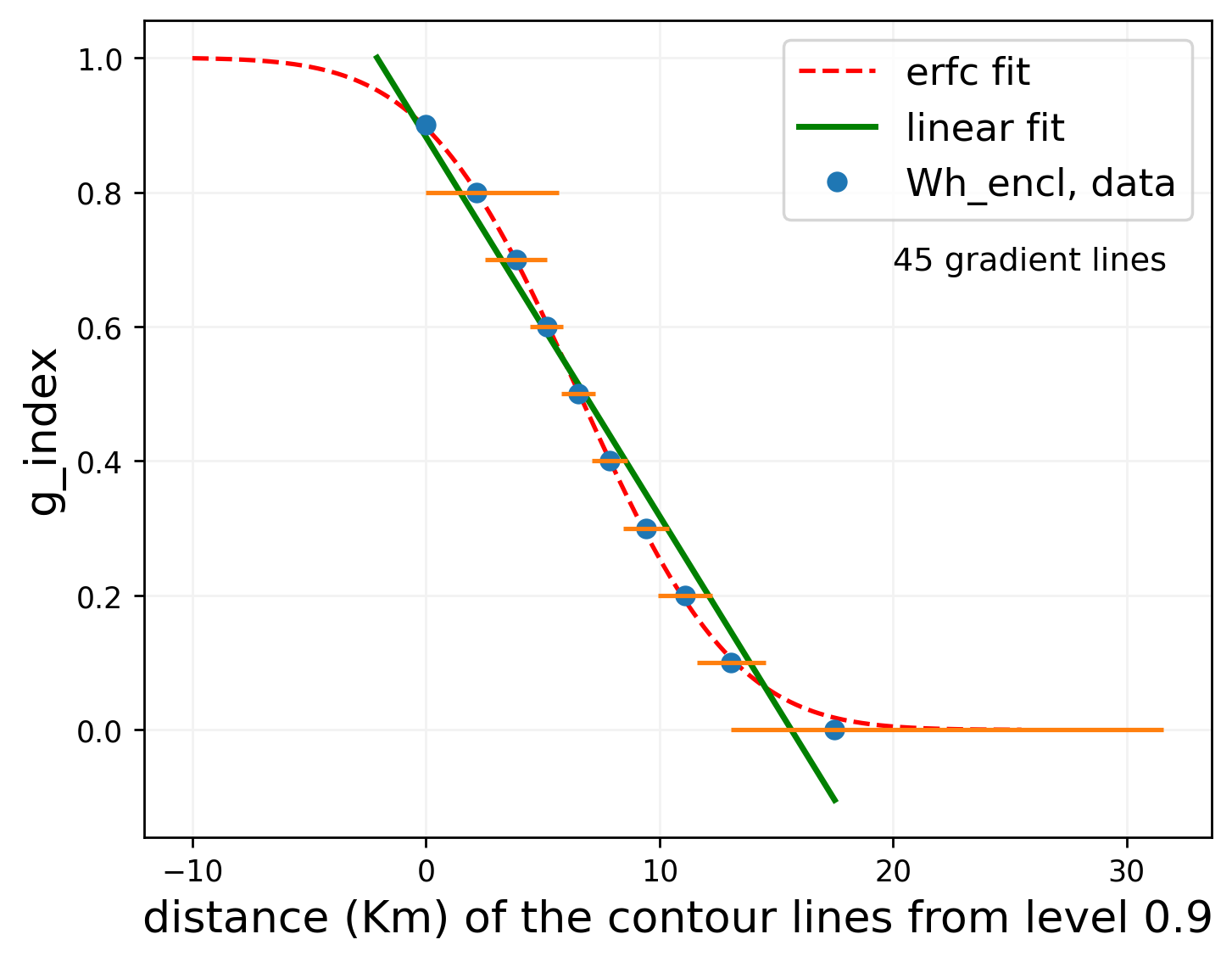}
                    \caption{Fit to {\it Wh\_encl}}
                    \label{Whe g_index_vs_gradient_line_distances_45_Lin.png}    
  \end{minipage}
\end{figure}
\newline
The data elaborated according to the definition of the distances $\avg{d^{(j)}}$ given in Eq.~(\ref{le distanze}) is shown in
Fig.~\ref{met g_index_vs_gradient_line_distances_45_Lin.png}\, for {\it meteoverbs} and in Fig.~\ref{Whe g_index_vs_gradient_line_distances_45_Lin.png}\, for 'Wh\_encl'; in both cases the fits to a straight line and to the erfc function are also reported: as already mentioned, these functions are both particular solutions of the diffusion equation, but corresponding to different diffusivities.
\newline
The reduced {\it chi-square} is a parameter used in statistics, which basically measures how close the experimental data are to a fit curve. In the case of {\it meteoverbs} the reduced chi-square for the linear fit is 0.00022, for the erfc is 0.00069, thus favoring evidently the linear fit; of the remaining features, only for the {\it Wh\_encl} case the erfc is definitely favorite, although for both fits the agreement between theoretical model and data is quite satisfactory.\newline 
\begin{figure}[!ht]
  \begin{minipage}{0.49\textwidth}
    \centering
    \includegraphics[width=\textwidth]
                    {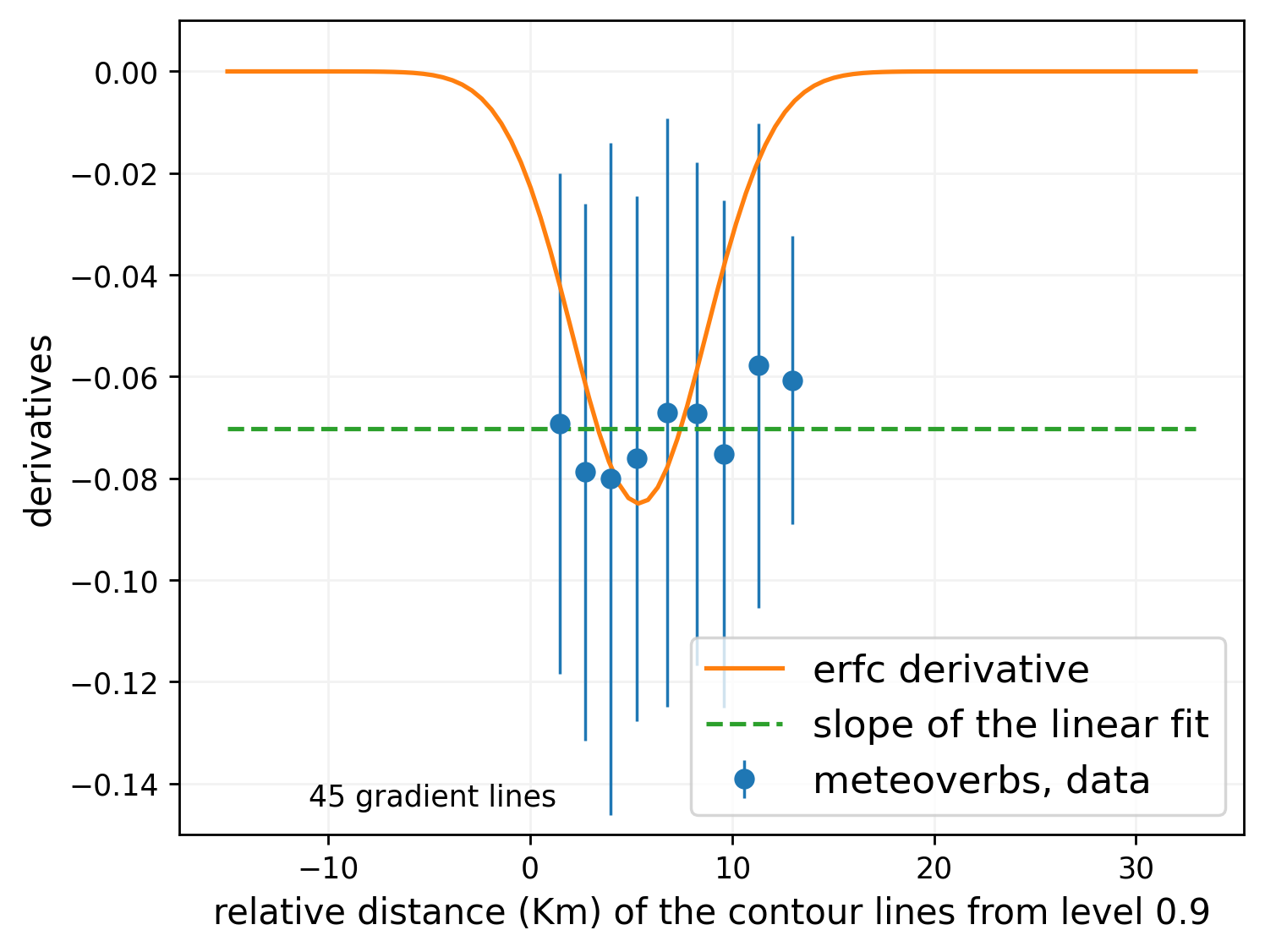}
                    \caption{Derivative of the erfc and the linear fits: {\it meteoverbs}}
                    \label{met der}
  \end{minipage}\hfill
  \begin{minipage}{0.49\textwidth}
    \centering
    \includegraphics[width=\linewidth]
                    {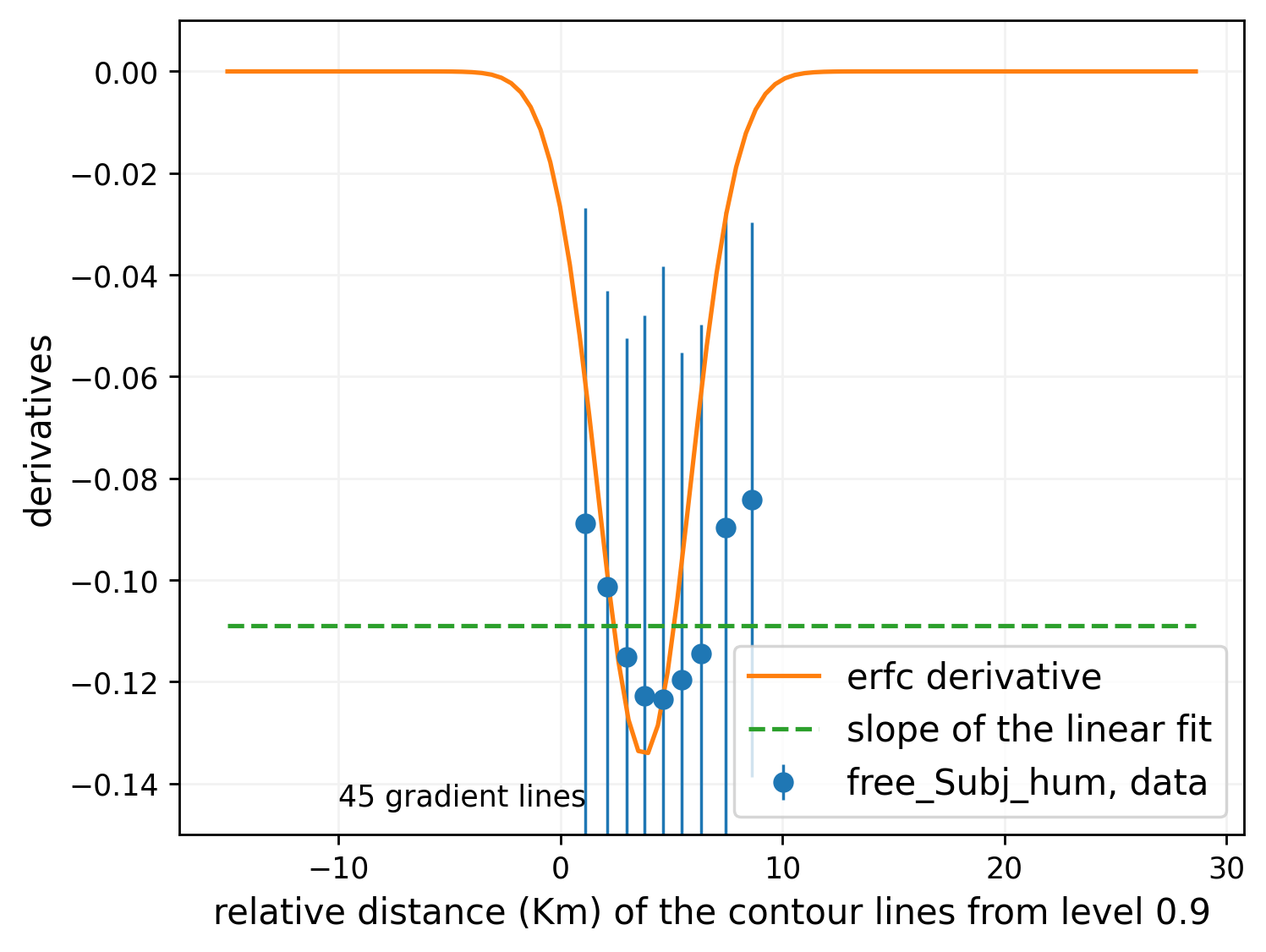}
                    \caption{Derivatives of the erfc and the linear fits: {\it free\_Subj\_hum}}
                    \label{fsh der}
  \end{minipage}
\end{figure}
\begin{figure}[!ht]
  \begin{minipage}{0.49\textwidth}
    \centering
    \includegraphics[width=\textwidth]
                    {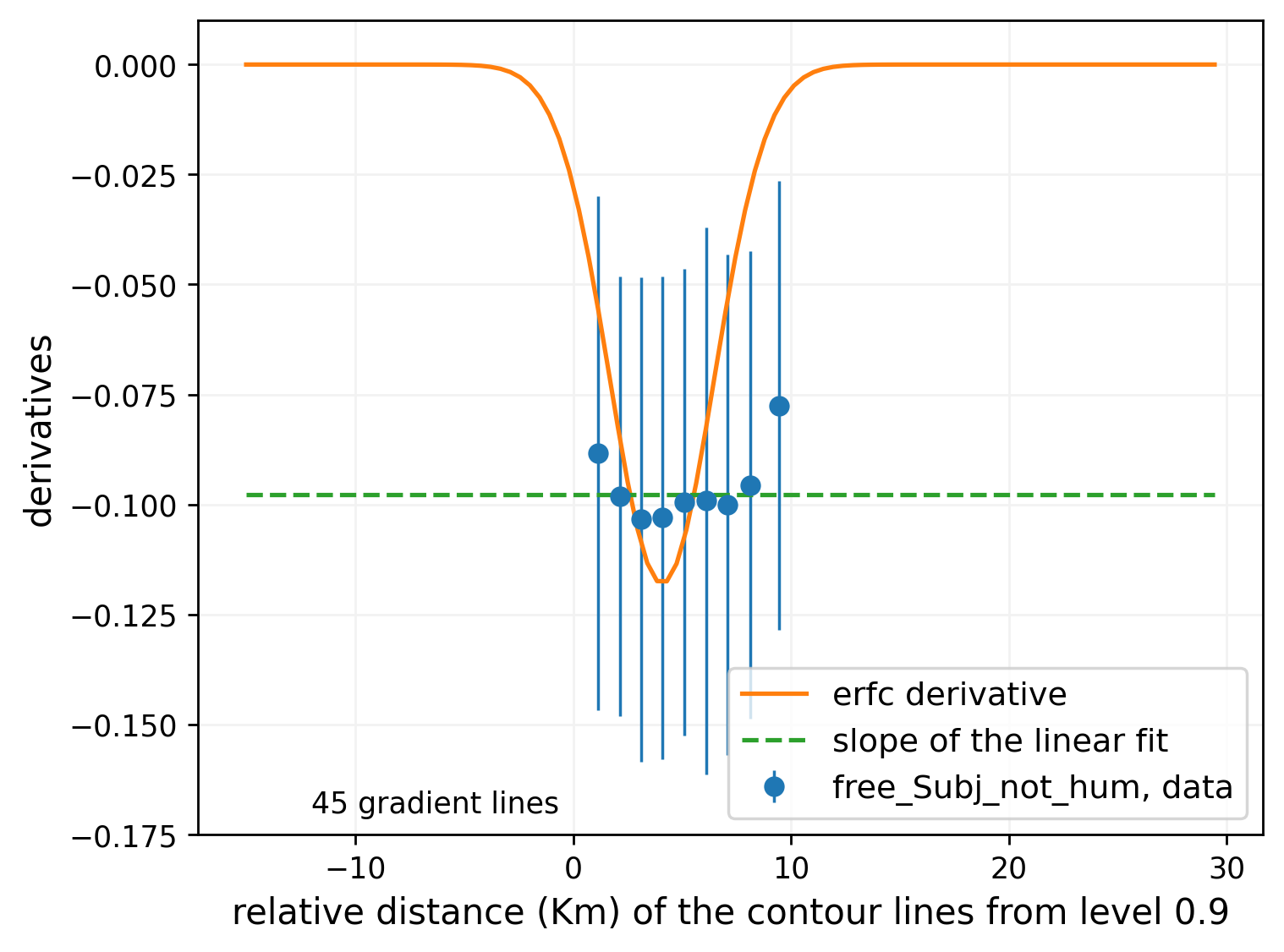}
                    \caption{Derivatives of the erfc and the linear fits: {\it free\_Subj\_not\_hum}}
                    \label{fsn der}
  \end{minipage}\hfill
  \begin{minipage}{0.49\textwidth}
    \centering
    \includegraphics[width=\linewidth]
                    {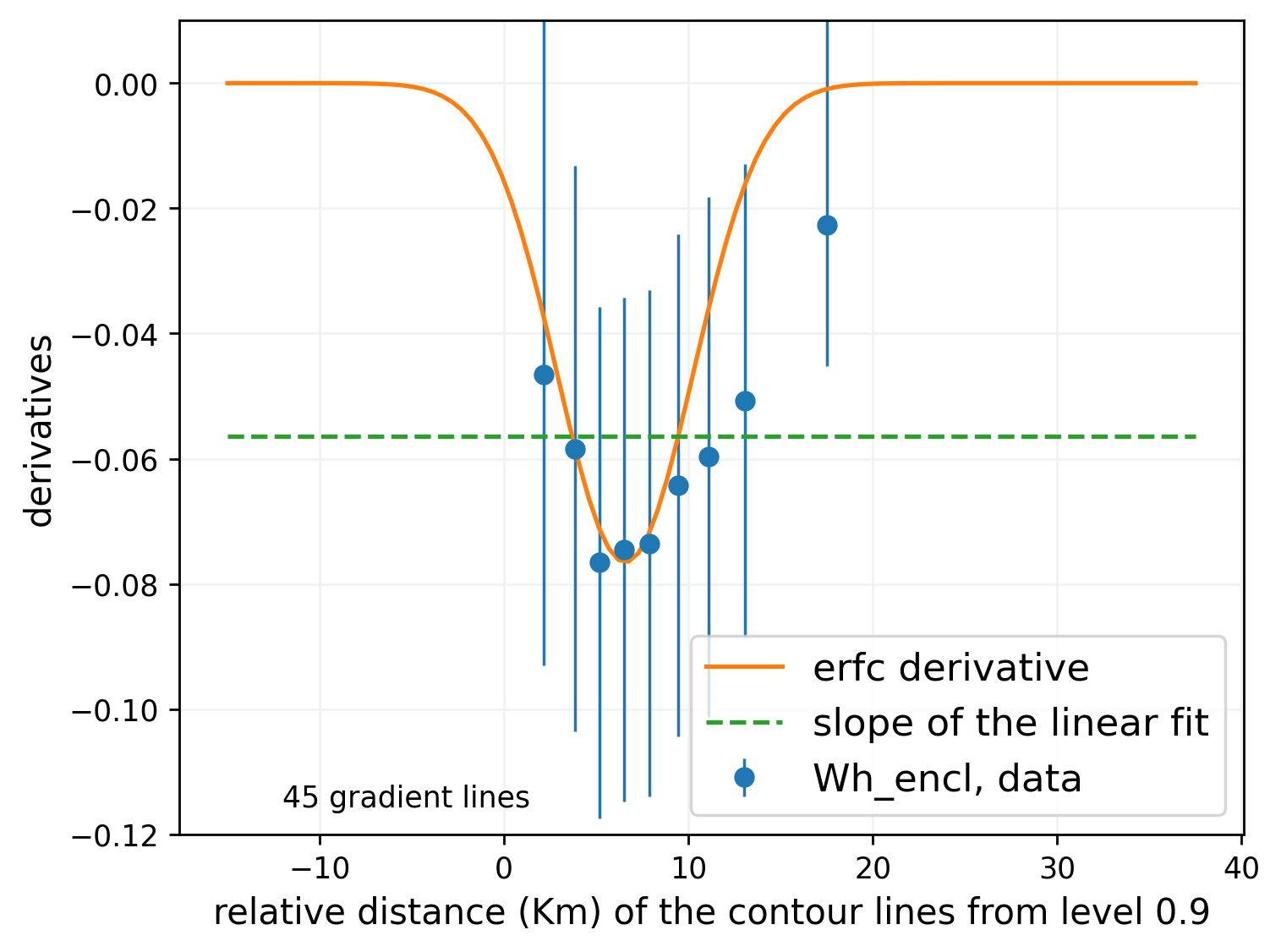}
                    \caption{Derivatives of the erfc and the linear fits: {\it Wh\_encl}}\label{Whe der}
  \end{minipage}
\end{figure}
\newline
In Fig~\ref{met der}, \ref{fsh der}, \ref{fsn der}, \ref{Whe der}\, the derivative of the fitting erfc is plotted against the data and the slope of the fitting straight line for each of the four features of the case study: it can be seen that the erfc is not at all out of play, confirming that the chi-squares are not so far apart, after all, a fact that is really intriguing and needs to be understood, because the linear solution and the erfc correspond to quite different diffusivities.\newline
More complete data are probably needed to settle the question, covering 100\% of the localities and also extending south of Rovigo, to include the whole of Emilia-Romagna. \newline
Table \ref{riassunto} presents a summary of the measured chi-squares. Two sets of gradient paths were tested, of 30 and 45 units respectively, from the 0.9 contour line to 0.0. Reported are the {\it chi-square} of each fit, the linear and the erfc. 
The last column reports the average width {\it w} in kilometers of the contour interval between the 0.9 and the 0.0 levels, which is indeed the \emph{gradient front}.
\newline
Following well known concepts in data analysis (see for instance \cite{Cowan}\,), marked in yellow, as having lower {\it chi-square}, is which fit is the favored one for each feature and for the 30 or 45 gradient path sets. 
\newline
\begin{table}[ht!]
    \begin{tabular}{||c | c | c | c | c ||}
      \hline
      &   & linear, & erfc, & w  \\
      feature & N & chisq  & chisq  & (Km) \\
      \hline\hline
      meteoverbs & 30 & \cellcolor[HTML]{FFFF00} 0.00050 & 0.00045 & 12\\
      \hline
      meteoverbs & 45 & \cellcolor[HTML]{FFFF00} 0.00021 & 0.00069  & 13\\
      \hline
      free\_Subj\_hum & 30 & \cellcolor[HTML]{FFFF00} 0.00022 & 0.00053 & 10 \\
      \hline
      free\_Subj\_hum & 45 & \cellcolor[HTML]{FFFF00} 0.00031 & 0.00041 & 9 \\
      \hline
      free\_Subj\_not\_hum & 30 & \cellcolor[HTML]{FFFF00} 0.00041 & \cellcolor[HTML]{FFFF00}0.00041 & 11 \\
      \hline
      free\_Subj\_not\_hum & 45 & \cellcolor[HTML]{FFFF00} 0.00009 & 0.00065 & 10 \\
      \hline
      Wh\_encl & 30 & 0.00201 & \cellcolor[HTML]{FFFF00} 0.00010 & 18 \\
      \hline
      Wh\_encl & 45 & 0.00295 & \cellcolor[HTML]{FFFF00} 0.00007 & 18 \\
      \hline
    \end{tabular}
  \caption{The chi-squares and the width (in Km) of the {\it gradient front}.}
  \label{riassunto}
\end{table}
\newline

\section{The time evolution}

The initial condition for the case study is that south of the Tyrol the whole territory was {\it i\_like} ($\mathcal{G} = 0$) until some time $t=0$; instead for the Tyrol itself $\mathcal{G} = 1$ permanently:
\begin{equation}
  \label{point initial conditions}
  \mathcal{G}(s, t) \,=\, 1\quad
  \forall s \le 0\,,\;\forall\: t \ge 0\,;
  \qquad
  \left.\mathcal{G}(s, t)\right\vert_{t=0} \,=\,0 \quad \forall\: s > 0\,.
\end{equation}
In principle these conditions would be sufficient to solve the given equation, were it not that the necessary {\it diffusivity function} is not known a priori. However, in the previous section for the given case study, data-driven solutions that fit the state at present were found, which call into play the result of \ref{furbata}, that a diffusivity function is uniquely fixed by the request that a given function is in fact a solution to the diffusion equation \ref{ord diff eq}.\newline
Needless to say, given diffusivity and a solution of the diffusion equation, one gets a feature flux according to Eq.~(\ref{Fick1})\,.
\newline
With reference to the brief historical note in subsection \ref{historic}\,, $\tau \approx 1000$\,yr will denote the time span since the beginning of the spread of Germanic syntactic features in North-Eastern Italy.
\subsection{The (piece-wise) linear solution}
\begin{figure}[!ht]
  \centering
  \includegraphics[width=0.9\textwidth]{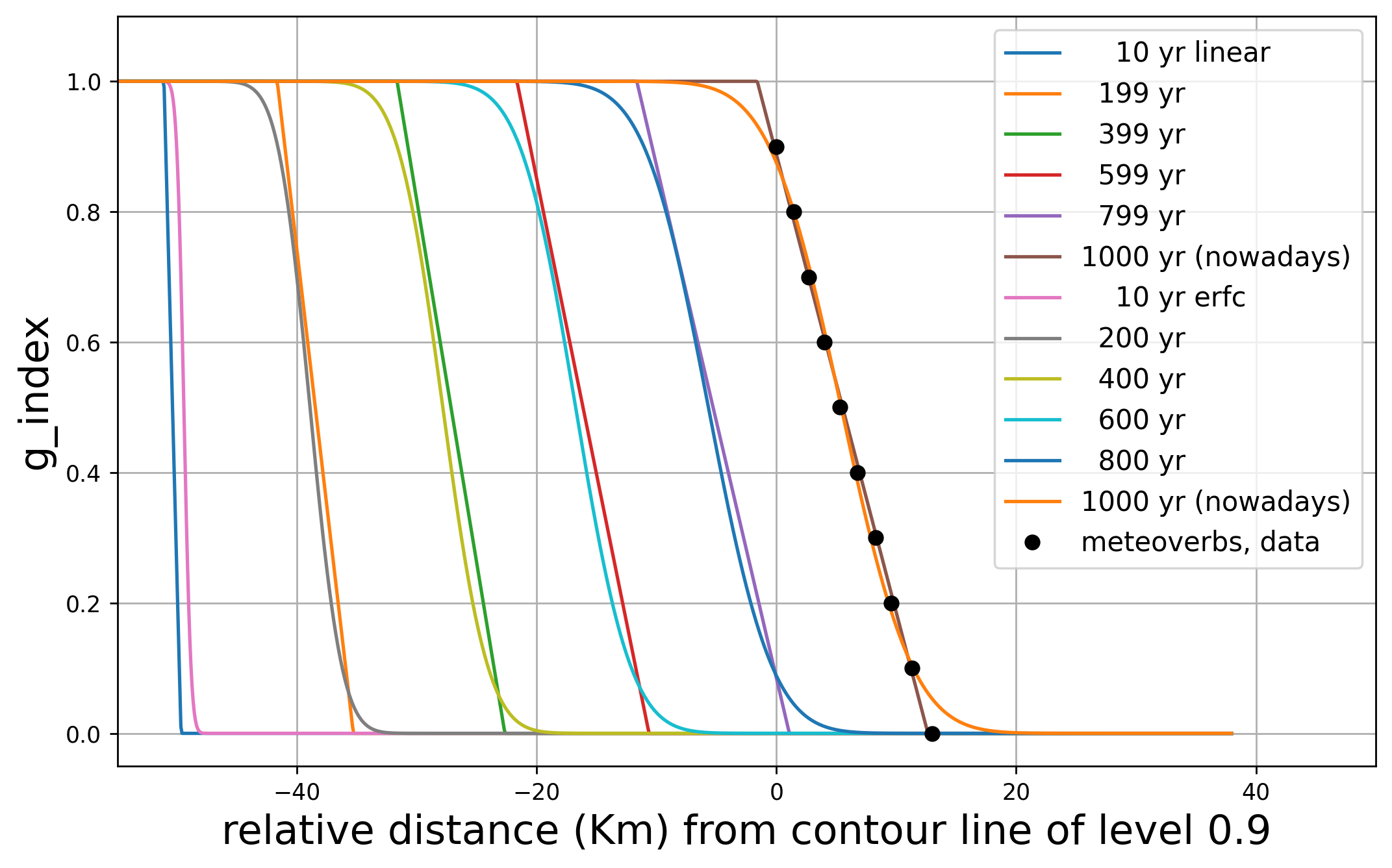}
  \caption{Time evolution of the linear and the erfc solutions, accounting for the {\it  convection} contribution as in Eq.~(\ref{simplest convection term})}
  \label{fig:all_evolution_over_time.png}
\end{figure}

Ignoring for a while the convection contribution, the linear solution of the diffusion equation is
\begin{equation*}
  \mathcal{G}_{\text{lin}}(s,t) \,=\, 1 - \chi\,\cfrac{s-s_1}{2\,\sqrt{t/\tau}}
  \,\equiv\, 1 \,-\, k(t)\,(s-s_1)\,,
\end{equation*}
where the angular coefficient depends on the time; namely
\begin{equation}
   k(t) \,=\, \cfrac{\chi}{2\,\sqrt{t/\tau}}\,.
   \label{k(t)}
\end{equation}
From the fit to the data one has
\begin{equation*}
  k(\tau) \,=\, \frac{\chi}{2} \,=\, 0.070\, (Km)^{-1}\,,\quad
  \tau \approx 1000\,\text{yr}.
\end{equation*}
Thus, ignoring the convection contribution, the point $s_1$\ such that $\mathcal{G}_{\text{lin}}(s_1,t) = 1$\,, is fixed in time: it marks the location of the point source of the linguistic diffusion; it is the starting point of the \emph{gradient front} as viewed in the downward direction.\newline
The point where the \emph{gradient front} ends at a given time $t$, that is the point $s_0(t)$ such that  $\mathcal{G}_{\text{lin}}(s_0(t),t)~=~0$, is given by
\begin{equation*}
  s_0(t) \,=\, s_1 + \frac{1}{k(t)} \,=\,
  s_1 + \frac{2}{\chi}\,\sqrt{\frac{t}{\tau}}\,.   
\end{equation*}
For $t=\tau$ one obtains the parameter $w$ shown in Table\ref{riassunto}, giving the span of the \emph{gradient front} at that time; namely (for \emph{meteoverbs})
\begin{equation*}
    w\,=\,s_0(\tau) - s_1  \,=\, \frac{1}{k(\tau)} \,\approx\, 14.28\,\text{Km}\,.   
\end{equation*}
The diffusivity is
\begin{equation}
  \eta(s,t) \,=\, C - \frac{(s-s_1)^2}{4\,t}\,,
\end{equation}
where $C$ is a constant to be fixed. At a time $t$, using \ref{k(t)}\, one has
\begin{equation}
  \eta(s_0(t),t) \,=\, C -\frac{(s_0(t)-s_1)^2}{4\,t} \,=\, C - \frac{1}{4\,t\,k(t)^2}
  \,=\, C - \frac{1}{\chi^2 \tau} \,,
\end{equation}
which is time independent. It follows that, by choosing
\begin{equation}
  C = \eta(s_0(t),t) = \frac{(s_0(t)-s_1)^2}{4\,t} = \frac{1}{\chi^2 \tau} \,,
\end{equation}
the diffusivity becomes a positive definite decreasing function for any time $t>0$ and
for $0 \le s \le s_0(t)$; null for $ s \ge s_0(t)$. Namely
\begin{subequations}
\begin{align}
  \label{eta for linear}
  &\eta(s,t) \,=\,\frac{(s_0(t)-s_1)^2}{4\,t}  - \frac{(s-s_1)^2}{4\,t}\qquad
  &&\forall t>0\,,\quad s_1 \le s \le s_0(t)\\
  &\eta(s,t) \,=\,0 &&\forall t>0\,,\quad s \ge s_0(t)\,.
\end{align}
\end{subequations}
Consequently, in the end, the following piece-wise linear solution is obtained:
\begin{subequations}
\begin{align}
  &\mathcal{G}_{\text{lin}}(s,t) \,=\, 1 - \chi\,\cfrac{s-s_1}{2\,\sqrt{t/\tau}}
  \qquad &&\forall t>0\,,\quad s_1 \le s \le s_0(t)\\
  &\mathcal{G}_{\text{lin}}(s,t) \,=\,0 &&\forall t>0\,,\quad s \ge s_0(t)\,.
    \label{no int. source linear}
\end{align}
\end{subequations}
\begin{figure}[!ht]
  \centering
  \includegraphics[width=0.65\textwidth]
                  {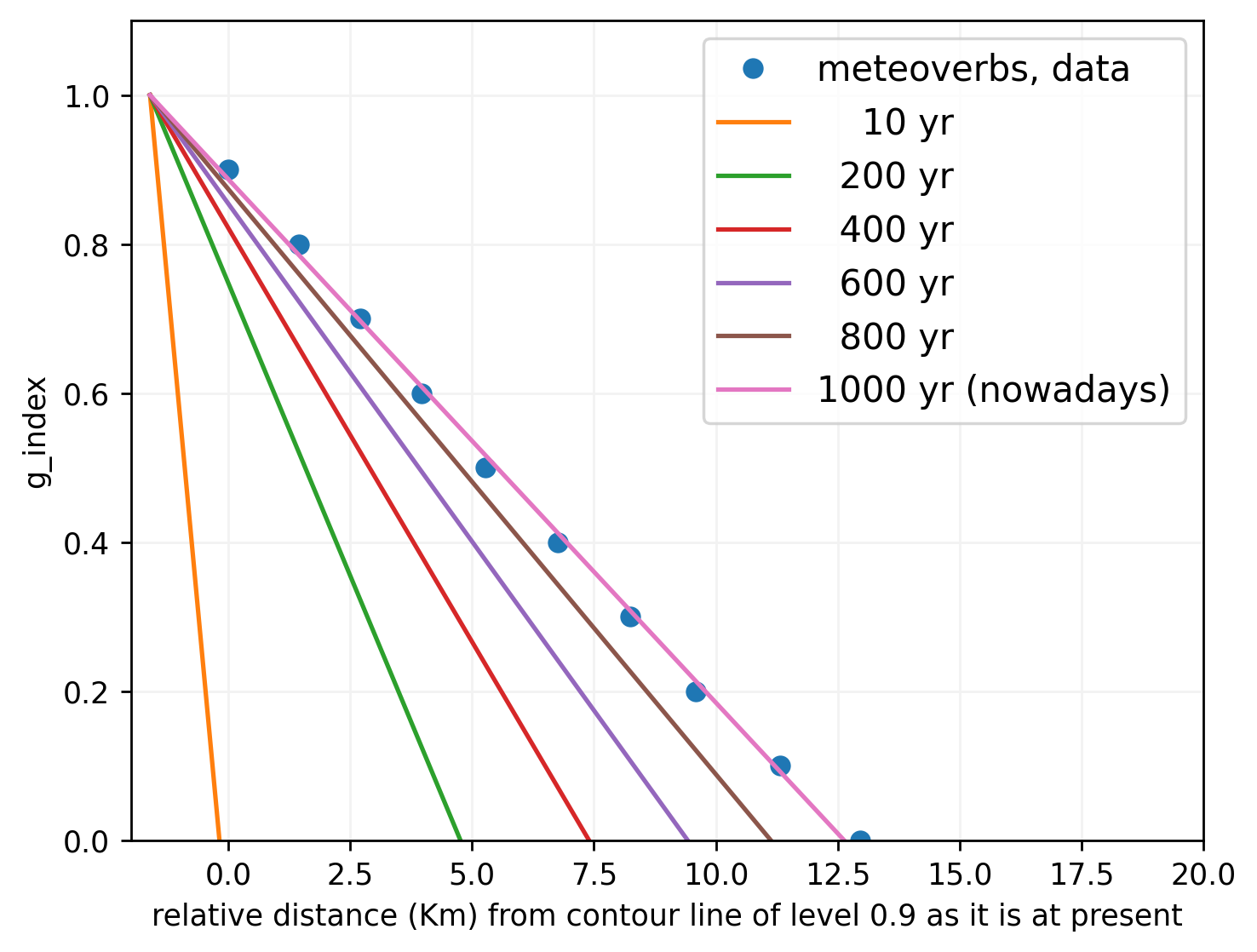 }
  \caption{Time evolution of the linear solution, no {\it convection term}.}
  \label{fig:no_internal_source_lin_evolution_over_time.png}
\end{figure}
\newline
Fig.~\ref{fig:no_internal_source_lin_evolution_over_time.png} shows the $\textit{g\_index}$ as a function of $s$ for {\it meteoverbs} at different times; for the remaining features one finds quite similar behaviors.
\newline
Accounting for the convection contribution like in Eq.~(\ref{one D diffu eq extended})\, (see (\ref{wave solution})\,), the piece-wise linear solution becomes
\begin{alignat}{2}
  \forall&\, t>0\,:  &&\nonumber\\
  &\mathcal{G}_{\text{lin}}(s,t) \,=\, 1  \quad  &&\text{for}\;s \le s_1(t)\nonumber\\
  &\mathcal{G}_{\text{lin}}(s,t) \,=\, 
  1 - \chi\,\cfrac{s-s_1-\lambda\,f(t/\tau) + \lambda\,f(1)}{2\,\sqrt{t/\tau}}
  \quad  &&\text{for}\; s_1(t) \le s \le s_0(t) \label{convection_linear}\\
  &\mathcal{G}_{\text{lin}}(s,t) \,=\,0 \quad &&\text{for}\; s \ge s_0(t)\nonumber\,.
\end{alignat}
This time, the location of the starting point of the \emph{gradient front}, as viewed in the downward direction, is a moving one, according to
\begin{equation}
  \label{init_front}
   s_1(t) = s_1 + \lambda\,f(t/\tau) - \lambda\,f(1)\,.
\end{equation}
Using the notation in Eq.~(\ref{k(t)})\,, one finds of course that the end point of the \emph{gradient front} moves faster, according to 
\begin{equation}
  \label{end_front}
   s_0(t) = s_1 + \lambda\,f(t/\tau) - \lambda\,f(1) \,+\, \frac{1}{k(t)}\,=\,
   s_1(t)\,+\, \frac{1}{k(t)}\,.
\end{equation}
\newline
For $f(t/\tau) = t/\tau$ and $\lambda= 50$ Km the plot of (\ref{convection_linear}) is shown in Fig.~\ref{fig:all_evolution_over_time.png} for \emph{meteoverbs} (together with the erfc solution to be analyzed later) at different times, roughly since year 1027 (see subsection \ref{historic}); the distances are relative to the the position today of the contour line of level 0.9\,.\newline
For the remaining features things go similarly, apart from different values of $w$ as shown in Table~\ref{riassunto}.

\subsection{The erfc solution}

The time evolving erfc solution is
\begin{equation}
  \mathcal{G}_{\text{erfc}}(s, t) \,=\,
  \frac{1}{2}\;
  \erfc\left(\Bigl[s-(s_0+\lambda)\,f(t/\tau)+\lambda\,f(1)\Bigr] \,\frac{\kappa}{2\,\sqrt{t/\tau}}\right)\,.
  \label{erfc_evol}
\end{equation}
Evaluated at $t=\tau$\ and $\lambda=0$\,, it is what was found to fit quite well the data, thus setting the constant parameters $\kappa$ and $s_0$.\newline
The diffusivity is a constant, parameterized as
\begin{equation}
  \eta\,=\, \frac{1}{\kappa^2\tau} \,\approx\, 0.011\; \text{Km${}^2$/yr}\,,
\end{equation}
evaluated using the fit values of  $\kappa$ and $s_0$.
The angular coefficient of the tangent in the inflection point is
\begin{equation}
   k(t) \,=\, \frac{\kappa}{2\,\sqrt{\pi\,t/\tau}}\,.
\end{equation}
Treating this in complete analogy with the linear solution, the value of $w$ for \emph{meteoverbs} (see Table~\ref{riassunto}) is found to be:
\begin{equation*}
    w\,=\,s_0(\tau) - s_1  \,=\, \frac{1}{k(\tau)} \,\approx\, 11.8\,\text{Km}\,.  
\end{equation*}
\begin{figure}[!ht]
  \centering
  \includegraphics[width=0.65\textwidth]{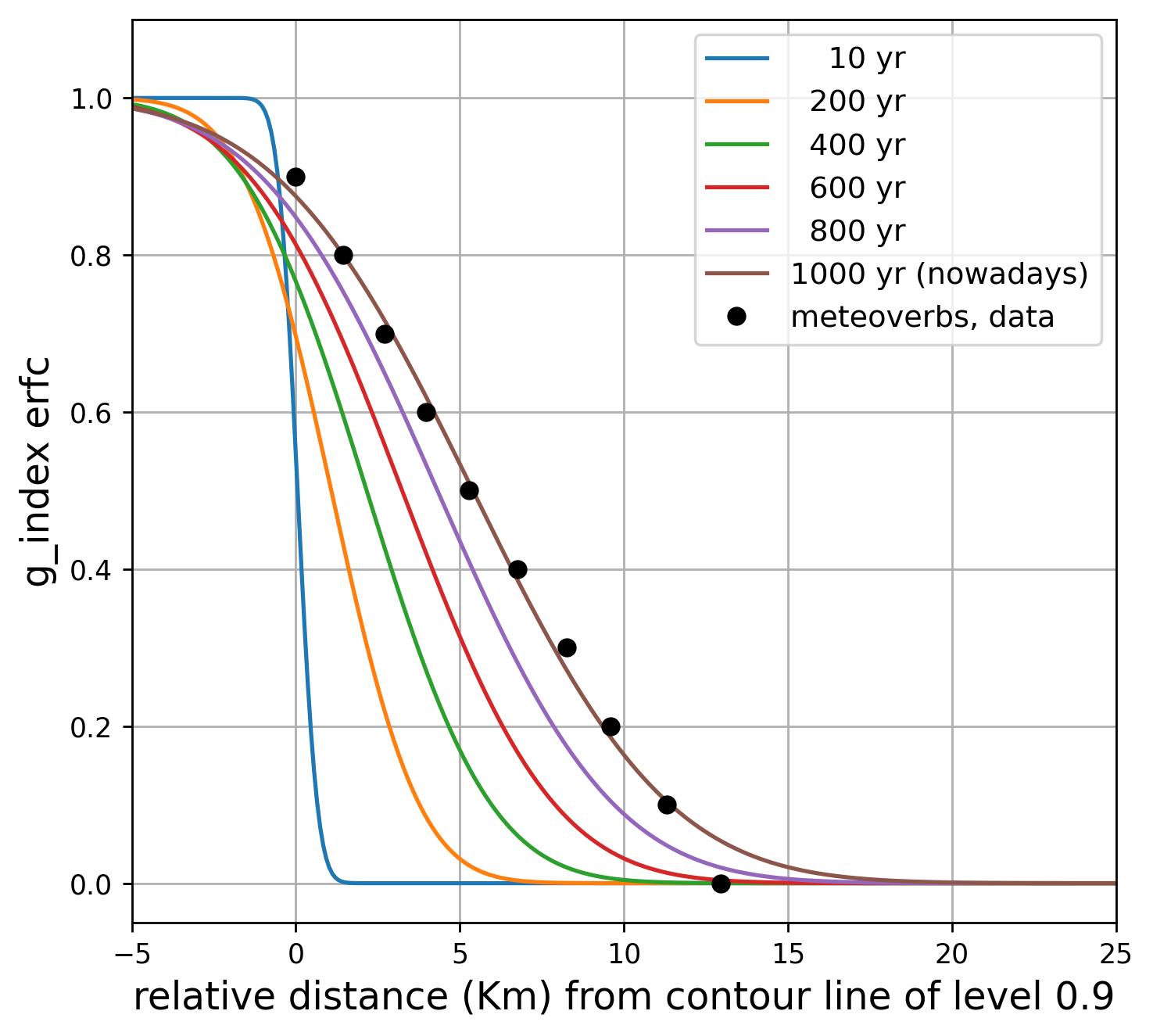}
  \caption{Time evolution of the erfc solution, not accounting for any {\it convection term}}
  \label{erfc_evolution_over_time_0.png}
\end{figure}
\newline
Imposing $\lambda=0$, of course the convection contribution is \emph{turned off}, so that the gradient front widens over time but, as a whole, does not progress geographically:
this is illustrated in Fig.~\ref{erfc_evolution_over_time_0.png} for meteoverbs; things go quite similarly for the remaining features.
\newline
Accounting for the \emph{convection} contribution with $\lambda=50$~Km and $f(t/\tau) = t/\tau$\,, for \emph{meteoverbs} the behaviour is shown in Fig.~\ref{fig:all_evolution_over_time.png}\,, confirming that the convection term has the effect of letting the \emph{gradient front} as a whole to advance over time.\newline

It is worthwhile and interesting to take a look at what happens when changing the function $f$ that determines the convection contribution. Here is an example (see \ref{wave solution}\,):
\begin{equation}
  f\left(t/\tau\right)\,=\,\frac{t}{\tau}\,
  \exp\left\{\frac{\tau}{\theta}\,\left(1\,-\,\frac{t}{\tau}\right)\right\}\,,
  \label{special}
\end{equation}
whose corresponding convection term is
\begin{equation}
  \frac{\lambda}{\tau}\,
  \left(1\,-\,\frac{t}{\theta}\right)\, \exp\left\{\frac{\tau}{\theta}\,\left(1\,-\,\frac{t}{\tau}\right)\right\}\,.
  \,\pdv{\mathcal{G}}{s}
  \label{special convection term}     
\end{equation}
\begin{figure}[!ht]
  \centering
  \includegraphics[width=0.9\textwidth]{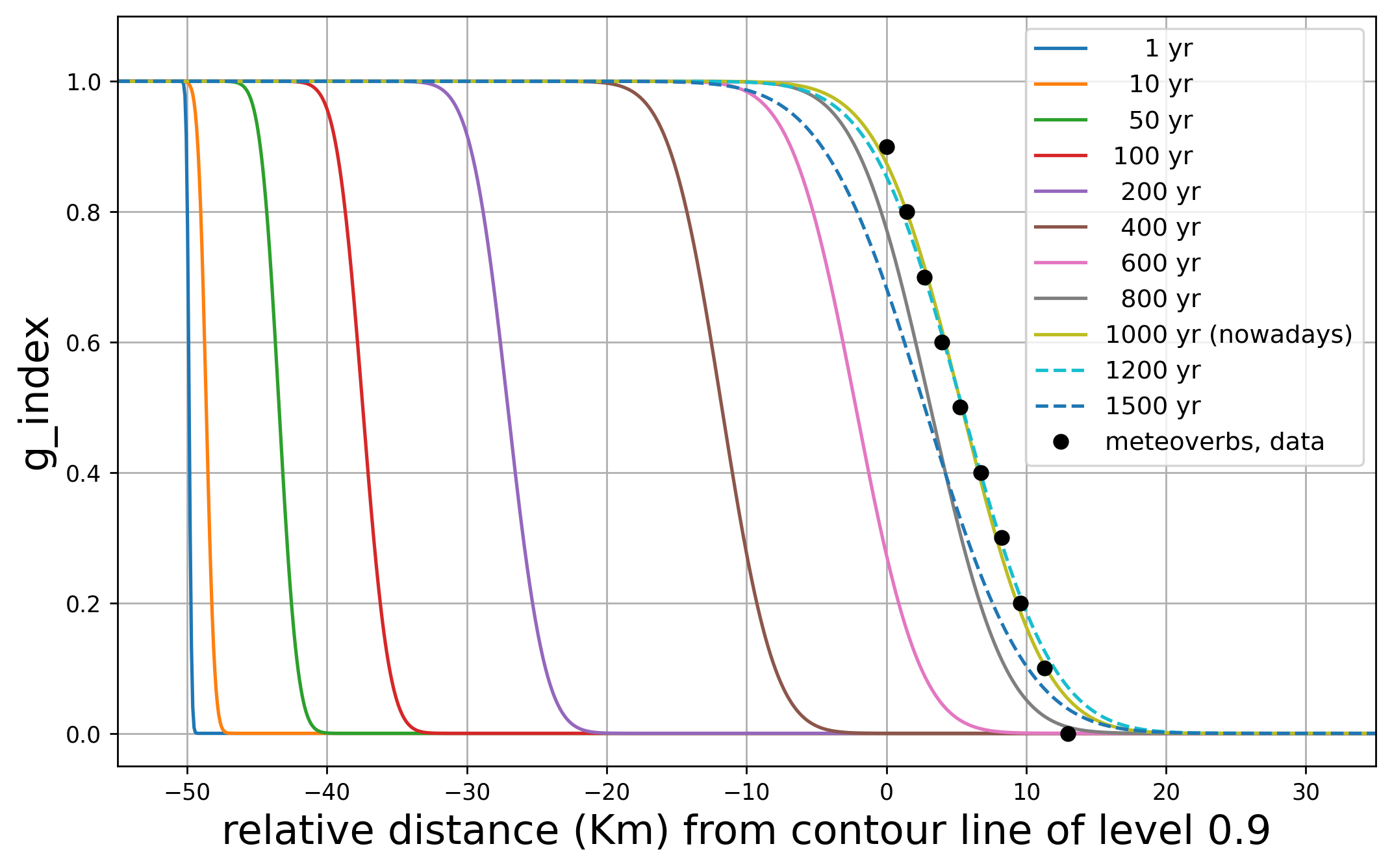}
  \caption{Time evolution of the linear solution, accounting for the {\it convection term} of Eq.~(\ref{special convection term})}
  \label{Erfc_special_evolution_over_time.png}
\end{figure}
\newline
Looking at its derivative, one finds that $f$ is an increasing function for $t < \theta$ and a decreasing one for $t > \theta$\,.
The effect is shown in Fig.~\ref{Erfc_special_evolution_over_time.png} for
$\theta = \text{year}\,1100$: at time $\theta$ the gradient front geographically stops advancing and starts to regress!

\section{The 'diffusion-convection' model in two dimensions and the Schmidt's 'waves'}

\begin{figure}[!ht]
  \centering
  \includegraphics[width=0.9\textwidth]{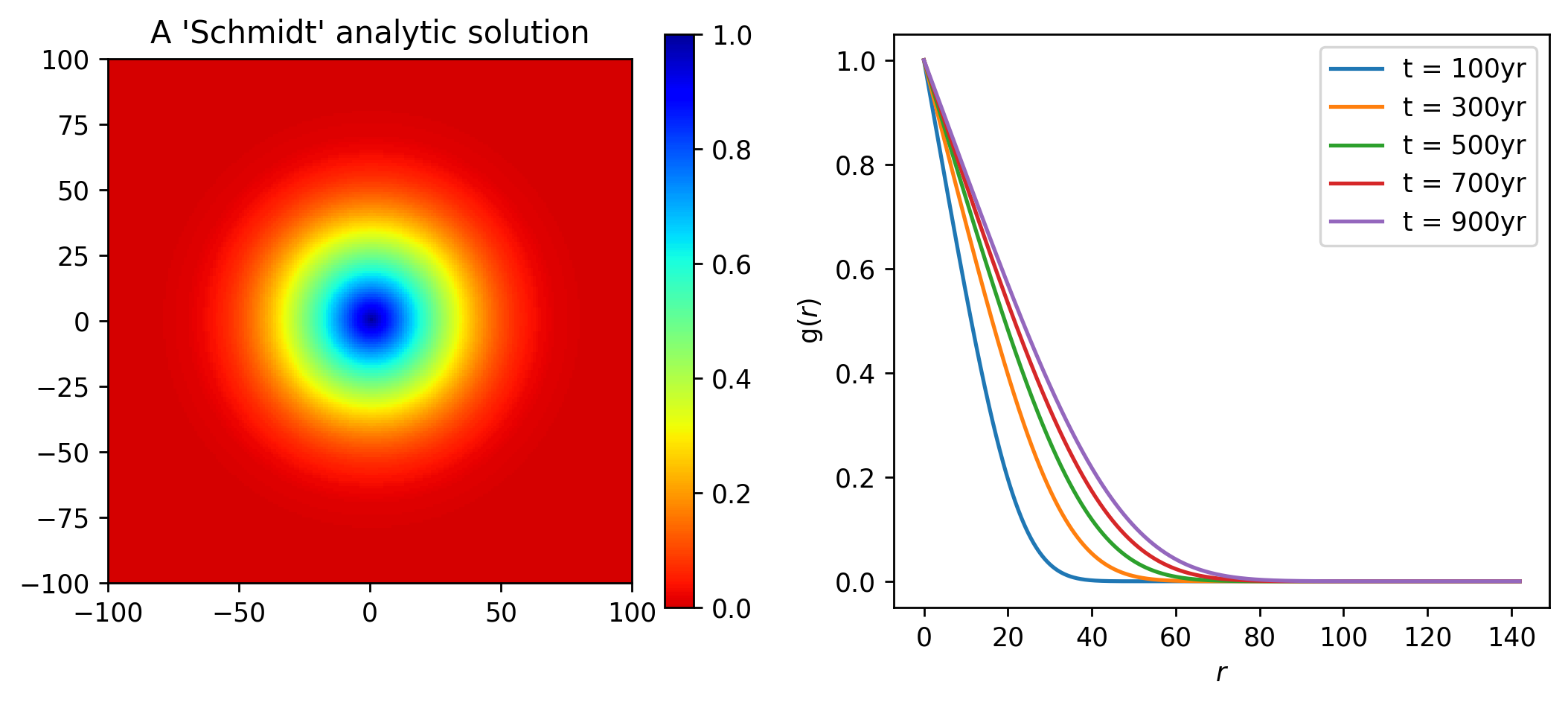}
  \caption{Heatmap view (left) and profiles of Schmidt type solution
    (Eq.~(\ref{Smidth sol})\,) at different times (right)}
  \label{Schmidt solution fig}
\end{figure}

\begin{figure}[!ht]
  \centering
  \includegraphics[width=0.9\textwidth]{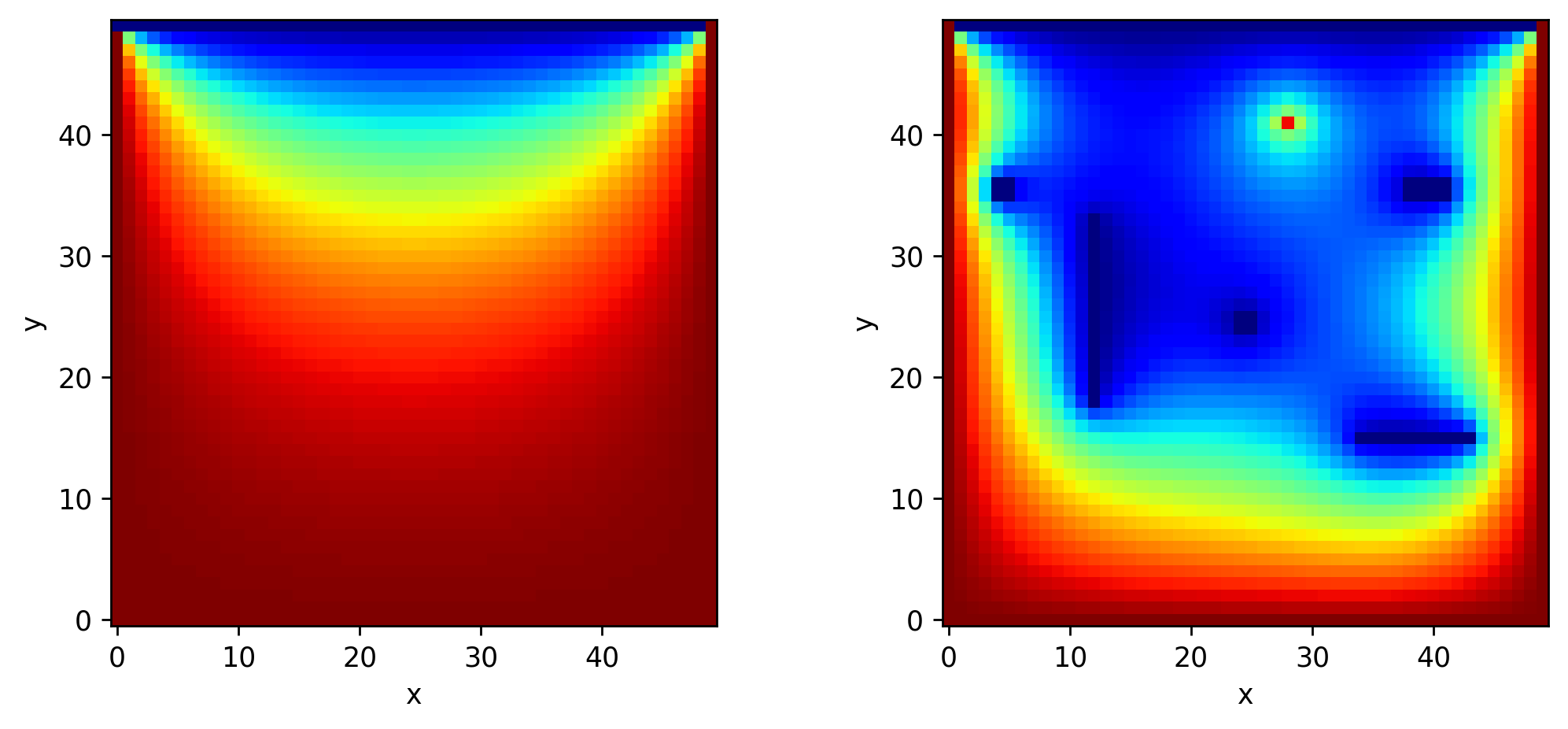}
  \caption{On the left a pure 'tidal' diffusion; on the right a combination of that tidal diffusion and Schmidt 'waves'}
  \label{Tide and Schmidt}
\end{figure}

In two dimension, using polar coordinates with usual notations, assuming for simplicity $\eta$ to be isotropic, the diffusion equation is written as
\begin{equation}
  \label{two D polar diffu eq}
  \pdv{\mathcal{G}}{t} \,=\,
  \pdv{}{r}\left(\eta\,\pdv{\mathcal{G}}{r}\right)\,+\,
  \frac{1}{r}\,\eta\,\pdv{\mathcal{G}}{r} \,.
\end{equation}
Taking
$$ \eta = \ho{\eta} = \text{const}\,, $$
there exists the so called fundamental solution (see for instance \cite{Crank}, pag.29) given by
\begin{equation}
  \label{fond sol two D}
  g(x,\,y,\,t) \,=\, \frac{1}{4\,\pi\,\ho{\eta}\,t}\;
  \exp{\left\{-\frac{x^2+y^2}{4\,\ho{\eta}\,t}\right\}}\,, 
\end{equation}
interpreted as the isotropic diffusion due to an instantaneous infinite source generated at time $t=0$ in the point $\vec{r}_0 = (0,\,0)$. The diffusion due to a whole line of such sources, all along the axis $y=0$\,, is
\begin{equation}
  \label{fond sol one D}
  G(s,\, t) \,=\, \frac{1}{4\,\pi\,\ho{\eta}\,t}\;
  \int_{-\infty}^{+\infty} \de{x}\,
  \exp{\left\{-\frac{x^2+s^2}{4\,\ho{\eta}\,t}\right\}} \,=\,
  \frac{1}{2\,\sqrt{\pi\,\ho{\eta}\,t}}\;
  \exp{\left\{-\frac{s^2}{4\,\ho{\eta}\,t}\right\}} \,, 
\end{equation}
which also is a particular solution of Eq.~(\ref{two D polar diffu eq})\,, representing a flux uniformly progressing orthogonally to the $y=0$ axis at distances $y\equiv s > 0$: it turns out, incidentally, to be the fundamental solution for the diffusion equation in one-dimension.
\newline
Following \cite{Crank}, pagg.13-14, one can consider the effect of additional, continuously distributed, instantaneous sources along the half-line $s \le 0$, getting
\begin{equation}
  G(s,\, t) \,=\, \frac{1}{2}\;\erfc{\left( \frac{s}{2\,\sqrt{\ho{\eta}\,t}}\right)}\,.
\end{equation}
This solution is exactly the same as in Eq.~(\ref{erfc})\,, but it has now a different interpretation: it comes from idealizing the Tyrol as a half-plane of uniformly distributed sources of linguistic innovation to the south, propagating like the sea onto the shore during the rising tide. It is an over-simplified picture, of course, and just serves illustrative purposes to capture the substance of things.
\newline
Of course, Fig.~\ref{erfc_evolution_over_time_0.png}, originally referring to the one-dimensional projection used in section \ref{The data analysis}, also illustrates the evolving profile of the \emph{gradient front} in this two-dimensions \emph{tidal model}, however over-simplified for using flat diffusivity and no convection component. In order for this tidal model to reach its potential for realism, it is necessary that the diffusivity well describes the effects that derive from the local history, from the people aptitude to incorporate linguistic innovations in their speech, even from the morphology of the territory: in such a case, Eq.~(\ref{diffu eq extended}) would certainly require the use of advanced numerical tools. 
\newline
\newline
In 1872 {\bf J.~Schmidt} \cite{Schmidt} formulated what he called the '{\bf wave theory}' of linguistic diffusion, proposing that a language feature from its region of origin spreads starting at some initial place, then affecting a gradually expanding cluster of dialects all around: '\emph{dialect diffusion spreads from given points of contact like waves generated by a stone fallen into the water}'.\\
Of course, strictly speaking, the 'wave' image was used quite improperly; however in the everyday speech people use, for instance, the term {\it heatwave} to mean incoming high temperatures, although physically heat propagates precisely by diffusion, not as a wave phenomenon.\newline
The expanding isotropic Schmidt's waves could be thought of in the meta\-phor of the diffusing dye, like circles of colour emanating continuously from a point \emph{source}: such source is nothing but the influence exerted in its home territory on the dialectal speech of fellow citizens by people who had acquired linguistic \emph{innovation} abroad, to call it as Schmidt used. As a matter of fact, coming to the case study, there are historical documents about contacts and presence of people and groups from North-Eastern Italy in Tyrol, due to economic and other interests, like notarial practices from Trentino and import to Venice of valuable wood for shipbuilding: all this may in fact have triggered Schmidt's waves at home places. \newline
It is possible to find analytically an example 'Schmidt solution' of
Eq.~(\ref{two D polar diffu eq})\,, using the following, plausible form of diffusivity: 
\begin{equation}
  \eta(r) \,=\, \ho{\eta}\, \frac{a}{r} \,,
\end{equation}
where $\ho{\eta}$ is a constant of diffusivity dimensions and $a$ is a length constant. The solutions can be found making the substitution
$$ z = \cfrac{r}{(3\,t)^{1/3}}\,,$$
so that the partial differential equation \ref{two D polar diffu eq} becomes an ordinary one, namely
\begin{equation}
  -\cfrac{z^2}{\ho{\eta}\,a}\;\odv{g}{z} \,=\,  \odv[order=2]{g}{z}\,.
\end{equation}
With passages analogous to those in \ref{error funct} one finds
\begin{equation}
  \label{Smidth sol}
  g(r,\,t) \,=\, 1\, -\, \cfrac{3}{\Gamma\left(\frac{1}{3}\right)}\;
  \bigintssss_0^{r/(3\,\ho{\eta}\,a\,t)^{1/3}}\de{x}\; e^{-x^3} \,,
\end{equation}
which is equal to $1$ in $r=0$ for $t>0$\,, while tends to zero for $r\rightarrow \infty$ and $t>0$ or for $r>0$ and $t\rightarrow 0^+$\,. Figure \ref{Schmidt solution fig} illustrate this. 
\newline
With the diffusivity depending on the coordinates only, Eq.~(\ref{two D polar diffu eq})  is linear, so that linear combinations of its solutions are themselves solutions. Of course this is true in particular for linear combinations of \emph{tidal-type} and Schmidt solutions, providing a new extremely powerful tool for trying realistic quantitative representations in two dimensions, and analysis of the diffusion of language features: it must be said for this model the same as for the diffusion-convection model, that almost certainly numerical techniques are to be used.
\newline
A very simple toy numerical simulation was carried out to compare a pure 'tidal' diffusion with a combination of 'tidal' and Schmidt's diffusion, the result of which is illustrated in Fig.~\ref{Tide and Schmidt}\,, where the upper edge symbolically represents the southern border of the Tyrol. What seems to be deduced from this numerical experiment is that Schmidt's 'waves' are perhaps even necessary to explain complex situations such as for the case study here: it is to be noted that it was possible to simulate even the persistence of an 'i-like island' rather close to the border of the Tyrol, like the real Ladin 'linguistic island'.
\newline
Constant diffusivity was assumed in the toy simulation, but even more realistic numerical simulations may be done, adapting the diffusivity to the geographic morphology of the territory and to the attitude to incorporate linguistic innovations in the local dialect: all the same as for the diffusion-convection model. It is worth mentioning that places and time of Schmidt's wave triggering would have to be subjected to documented historical facts. As soon as the completion of the survey on the dialects of North-Eastern Italy is available and pertinent historical documents have been collected, the tools prepared here will be used for an exhaustive study.
\newline
The simulations were done in Python; code samples can be found for example in \cite{Nervadof}\,.
\section{Conclusions}
Using Geographic Data Science concepts and techniques, a mathematically analyzable representation of the diffusion of linguistic features was introduced. This was done in particular for the case study of the Germanic language structures in dialects of the North-Eastern Italy, using the diffusion-convection equation, well known in physics and rephrased in the linguistic studies context. Despite simplifications and approximations, convincing pictures of time evolution could be deduced for the case study, assuming historically founded initial conditions and constraining solutions of the diffusion-convection equation to fit the available data.\newline
Most importantly, it was possible to establish that the so called Schmidt's diffusion 'waves' are among the solutions of the diffusion equation and that complex real-life situations could be simulated and studied by superposing pure diffusion and Schmidt 'waves' solutions. This approach and the diffusion-convection model should be regarded as competitive with each other. Once both are provided with appropriate local diffusivity hypothesis and initial and boundary conditions, they will allow, on the one hand, to approach the question of language contact from a completely different perspective than that currently prevailing in linguistics and, on the other hand, to model the historical colonization of the higher Alpine regions by German immigrants in the Middle Ages, in a way that is substantially different from and complementary to those used in history and sociology. Indeed, hopefully a novel approach to linguisics as a science has been proposed, to be applied to many different cases then the one taken here as an example.

\section*{Acknowledgment}
The author is very pleased to thank Ermenegildo Bidese of the Department of Humanities of the University of Trento for his invitation to consider the subject of metrical dialectology and for the subsequent collaboration, the result of which is the present work.

%
%
\appendix
\section{Some Maths}
\subsection{The one dimension approximation of the diffusion equation}
\label{diffu to one dim}
By its very definition the gradient of a function is in fact the directional derivative along a gradient line. Denoting by $\vec{\gamma}:\Rbb \rightarrow \cal{M}$ a gradient line over the (geographic) map $\cal{M}$, by $\vec{u}(s)$ the tangent versor to $\vec{\gamma}$ in $s$ and with $D_{\vec{u}(s)}$ the directional derivative along $\vec{u}(s)$, one has
\begin{equation}
  \vec{\nabla} \mathfrak{G}(\gamma(s)) \,=\,
  \vec{u}(s)\, D_{\vec{u}(s)} \mathfrak{G} (\vec{\gamma}(s)) =
  \vec{u}(s)\, (\frac{d}{ds} \mathfrak{G} \circ \vec{\gamma})(s) \,.
\end{equation}
The parameter $s \in \Rbb$ will be distance (in Km) along $\vec{\gamma}$ from a start point on the contour line of level 0.9. From Eq.~(\ref{diffu eq}) one obtains
\begin{equation}
  \label{along a path}
  - \,(\eta\circ\vec{\gamma})(s)\;
  \frac{d}{ds} (\mathfrak{G} \circ \gamma)(s) \,=\,
  \vec{u}\,\cdot\, \Phi(\vec{\gamma}(s))\,=\,
  (\left|\vec{\Phi}\right|\circ\vec{\gamma})(s) \,.
\end{equation}
Introducing
\begin{equation}
  \label{new notationsl}
  \mathcal{G}_\gamma(s)  \defeq  (\mathfrak{G} \circ \vec{\gamma})(s)\,,
  \quad
  \varphi_\gamma(s) \defeq  (\abs{\vec{\Phi}}\circ\vec{\gamma})(s)\,, \quad
  \eta_\gamma(s) \defeq \left(\eta\circ\vec{\gamma}\right)(s)\,,
\end{equation}
Eq.~(\ref{along a path}) is written as
\begin{equation}
  \label{Fick3}
  - \eta_\gamma\,\frac{d\mathcal{G}_\gamma}{ds}(s)\, =\, \varphi_\gamma(s)\,,
\end{equation}
and Eq.~(\ref{diffu eq}) as
\begin{equation}
  \pdv{\mathfrak{G}_\gamma}{t}(s) =
  \vec{\nabla} \cdot \left(\eta_\gamma(s)\, \vec{u}(s)\,
  \odv{\mathfrak{G}_\gamma}{s}(s) \right) \,.
\end{equation}
Assuming $\vec{\gamma}$ almost a straight line, that is $\vec{u}$ almost a constant, one has
\begin{align}
  \label{the diffu in 1 D}
  \pdv{\mathfrak{G}_\gamma}{t}(s) &\,=\,
  \vec{u}(s)\cdot \vec{\nabla} 
  \left(\eta_\gamma(s)\,\odv{\mathfrak{G}_\gamma}{s}(s)\right)\, +\,
  \eta_\gamma(s)\,\odv{\mathfrak{G}_\gamma}{s}(s)\;(\vec{\nabla} \cdot \vec{u})\\
  &\,\approx\,
  \vec{u}(s)\cdot \vec{\nabla} 
  \left(\eta_\gamma(s)\,\odv{\mathfrak{G}_\gamma}{s}(s)\right)\, =\,
  \frac{d}{ds}\left(\eta_\gamma(s)\,\odv{\mathfrak{G}_\gamma}{s}(s)\right)\,.
\end{align}
\vspace*{5mm}
\subsection{The error function complement}
\label{error funct}
From Eq.~(\ref{ord diff eq})\,, with $\eta$ constant, one has
\begin{equation*}
  -\,2\,\frac{z}{\eta}\,\odv{\mathcal{G}}{z} \,=\,
  \odv[order={2}]{\mathcal{G}}{z} 
\end{equation*}
and consequently
\begin{equation*}
  -\,2\,\zeta\,\odv{\mathcal{G}}{\zeta} \,=\,
  \odv[order={2}]{\mathcal{G}}{\zeta}\,, \qquad \zeta = \frac{s}{2\,\sqrt{\eta\,t}}\,,
\end{equation*}
or
\begin{equation*}
  -\,2\,\zeta \,=\,
  \odv{}{\zeta}\left(\ln\left(\odv{\mathcal{G}}{\zeta}\right)\right)\,,
\end{equation*}
hence
\begin{equation*}
  \odv{\mathcal{G}}{\zeta} \,=\, \mathcal{C}\, e^{-\zeta^2}\,,
\end{equation*}
thus
\begin{equation*}
  \mathcal{G}(s,\,t) \,=\,
  A \,+\, \mathcal{C}\,\int_0^{\;s/(2\,\sqrt{\eta\,t})} \de{u}\; e^{-u^2} \,.
\end{equation*}
Imposing
\begin{align*}
  1 &\,=\, \lim_{s\rightarrow -\infty}\mathcal{G}(s,\,t\neq 0) \,=\,
  A - \mathcal{C}\,\frac{\sqrt{\pi}}{2}\,,\\
  0 &\,=\, \lim_{s\rightarrow +\infty}\mathcal{G}(s,\,t\neq 0) \,=\,
  A + \mathcal{C}\,\frac{\sqrt{\pi}}{2}\,,\\
\end{align*}
one gets
\begin{equation*}
  \mathcal{G}(s,\,t) \,=\, \frac{1}{2}\, 
  \left(1 \,-\,\frac{2}{\sqrt{\pi}}\,\int_0^{\;s/(2\,\sqrt{\eta\,t})}\de{u}\;e^{-u^2}\right)
  \,\equiv\, \frac{1}{2}\,\erfc{\left( \frac{s}{2\,\sqrt{\eta\,t}}\right)} \,,
\end{equation*}
thus proportional to the function known in literature as {\it error function complement} ($\erfc$). \newline
It is to be noted that for $s>0$ as $t\rightarrow 0^+$ this function approximate more and more the step Heaviside distribution.\newline
Notice that $\mathcal{G}(0,\,t\neq0) = 1/2$\,.
\vspace*{5mm}
\subsection{ Obtaining solutions of the diffusion-convection equation from solutions of the pure diffusion equation}
\label{wave solution}
Let
\begin{equation}
  -\,2\,z\,\odv{g}{z} \,=\,
  \odv{}{z}\left( \eta\,\odv{g}{z} \right)\,.
  \label{presupposto}
\end{equation}
If
\begin{equation*}
  g=g\left(\frac{s-s_0}{2\sqrt{t}}\right) \qquad \text{and} \quad
  z\equiv \frac{s-s_0}{2\sqrt{t}}\,,
\end{equation*}
using the chain rule of the derivatives, one has
\begin{equation*}
  \pdv{g}{t} \,=\, \pdv{}{t}\left(\frac{s -s_0}{2\sqrt{t}}\right)\;
  \odv{g}{z}\,=\,
  \frac{1}{4t} \left(-2z\,\odv{g}{z}\right) \,=\,
  \frac{1}{4t}\, \odv{}{z}\left( \eta\,\odv{g}{z} \right) \,=\,
  \pdv{}{s}\left(\eta\,\pdv{g}{s}\right)\,.
\end{equation*}
If instead
\begin{equation}
  g = g\left(\frac{s -s_0 - \lambda\,f(t/\tau)}{2\sqrt{t}}\right)
  \qquad \text{and} \quad
  z\equiv \, \frac{s -s_0 - \lambda\,f(t/\tau)}{2\sqrt{t}} \,, 
  \label{general gradient front progressing}
\end{equation}
where $\lambda$ and $\tau$ are respectively a length and a time constant parameters, one has
\begin{align*}
  \pdv{g}{t} &\,=\, \pdv{}{t}\left(\frac{s -s_0 -\lambda \,f(t/\tau)}{2\sqrt{t}}\right)\;
  \odv{g}{z}\,=\,
  \pdv{}{t}\left(\frac{s -s_0}{2\sqrt{t}}-\frac{\lambda}{2\,\sqrt{t}}\,f(t/\tau)\right) \;
  \odv{g}{z} \\
  &\,=\,\left( - \frac{s -s_0}{4t\sqrt{t}} \right) \;\odv{g}{z} \,+\,
  \frac{\lambda}{4\,t\,\sqrt{t}}\,f(t/\tau) \;\odv{g}{z}\,-\,
  \frac{\lambda}{\tau}\,\frac{1}{2\,\sqrt{t}}
  \,\left.\odv{f(t')}{t'}\right|_{t'= t/\tau}\,\;\odv{g}{z} \\
  &\,=\,
  \left( - \frac{s -s_0 - \lambda\,f(t/\tau)}{4t\sqrt{t}} \right) \;\odv{g}{z}
  \,-\, \frac{\lambda}{\tau}\,
  \left.\odv{f(t')}{t'}\right|_{t'= t/\tau}\,\;\,\frac{1}{2\,\sqrt{t}}\,\odv{g}{z} \\
  &\,=\,
  \frac{1}{4t}\left(-2z\,\odv{g}{z}\right) \,-\,
  \frac{\lambda}{\tau}\left.\odv{f(t')}{t'}\right|_{t'= t/\tau}\,\pdv{g}{s}\\
  &\,=\,\frac{1}{4t}\odv{}{z}\left( \eta\,\odv{g}{z}\right) \,-\,
  \frac{\lambda}{\tau}\left.\odv{f(t')}{t'}\right|_{t'= t/\tau}\,\pdv{g}{s}\\ 
  &\,=\, \pdv{}{s}\left(\eta\,\pdv{g}{s}\right) \,-\,
  \frac{\lambda}{\tau}\left.\odv{f(t')}{t'}\right|_{t'= t/\tau}\,\pdv{g}{s}\,.
\end{align*}
Thus
\begin{equation}
  \pdv{g}{t} \,=\,
  \pdv{}{s}\left(\eta\,\pdv{g}{s}\right) \,-\,
  \frac{\lambda}{\tau}\,\left.\odv{f(t')}{t'}\right|_{t'= t/\tau}\,\pdv{g}{s}\,,
  \label{general convection term}
\end{equation}
Taking for example
\begin{equation*}
  f\left(\frac{t}{\tau}\right) = \frac{t}{\tau}
\end{equation*}
\ref{general gradient front progressing} becomes
\begin{equation}
  g = g\left(\frac{s -s_0 - (\lambda/\tau)\,t}{2\,\sqrt{t}}\right)\,,
  \label{linear progressing}
\end{equation}
with {\it convection term} simply
\begin{equation}
  v\,\pdv{g}{s}\,.
  \label{simplest convection term}
\end{equation}
In this case the {\it gradient front} progresses at the constant speed $v= \lambda/\tau$\,.\newline
As a different, instructive example, one sets
\begin{equation}
  f\left(t/\tau\right)\,=\,\frac{t}{\tau}\,
  \exp\left\{\frac{\tau}{\theta}\,\left(1\,-\,\frac{t}{\tau}\right)\right\}
\end{equation}
for which the {\it convection term} changes to
\begin{equation}
  \frac{\lambda}{\tau}\,
  \left(1\,-\,\frac{t}{\theta}\right)\,
  \exp\left\{\frac{\tau}{\theta}\,\left(1\,-\,\frac{t}{\tau}\right)\right\}\,.
\end{equation}
In this second case, for $t\ll\tau$\,, $f$ has approximately has approximately the same growth as in the previous choice, however bending to flatten until $t= \theta\; (\theta > \tau)$, where it begins to decrease: in view of \ref{general gradient front progressing} this means that the {\it gradient front} initially progresses at a speed approximately constant, then slows down, stops and finally slowly begins to regress.

\vspace*{5mm}
\subsection{{Searching the {\it diffusivity} function for data driven shapes.}}
\label{furbata}
As shown above, the data suggest some possible functional laws of 
$\mathcal{G}(z)$. The problem is to check whether there is a diffusivity function that allows the Eq.~(\ref{ord diff eq}) to have such function as a solution.\newline
Taking indeed
\begin{align}
  \label{a1}
  \eta(z ; \mathcal{G}(z)) \,=\, 2\;\left(\odv{\mathcal{G}(z)}{z}\right)^{-1}\,
  \int_0^z \!\de{\xi}\,\xi\,\odv{\mathcal{G}(\xi)}{\xi}\,,
\end{align}
in the domain of $z$ where the above is well defined, it follows
\begin{align*}
  &\odv{}{z}\left( 
  \left[2\;\left(\odv{\mathcal{G}(z)}{z}\right)^{-1}\,\int_0^z \!\de{\xi}\,\xi\, \odv{\mathcal{G}(\xi)}{\xi}\right]
  \,\left(\odv{\mathcal{G}(z)}{z}\right)\right)
  \,=\,\,2\,\odv{}{z}\left( \int_0^z \!\de{x}\,x\,\cfrac{\de{\mathcal{G}(z)}}{\de{x}} \right)\\
  \,=\,&2\,z\, \cfrac{\de{\mathcal{G}(z)}}{\de{z}} \,,
\end{align*}
because in the first term of the chain of equalities $(d\mathcal{G}/dz)^{-1}$ and $d\mathcal{G}/dz$ elide each other.
%
%
%
\vspace*{5mm}

\section*{Data availability statement}
The raw data are freely available at the Eurac Research CLARIN Centre (ERCC) (http://hdl.handle.net/20.500.12124/46).
\vspace*{5mm}




\begin{thebibliography}{00}

\bibitem{Halliday}
  M.A.K.~Halliday, J.~Webster,
  On Language and Linguistics. Continuum International Publishing Group. p. vii,
  (2006),
  ISBN 978-0-8264-8824-4.

\bibitem{Martinet}
  A.~Martinet,
  Elements of General Linguistics. Studies in General Linguistics,
  London: Faber (1960).

\bibitem{Labov 2}
  W.~Labov,
  Transmission and Diffusion,
  Language 83(2):344–387 (2007).

\bibitem{Padoan}
  A.~Padovan, A.~Tomaselli, M.~Bergstra, N.~Corver, R.Etxepare, S.Dold,
  Minority languages in language contact situations: three case studies on language change,
  Us Wurk 65(3/4):146–174 (2016).

\bibitem{Schmidt}
  J.~Schmidt,  
  'Die Verwandtsch\"{a}ftsverh\"{a}ltnisse der indogermanischen Sprachen'
  Weimar: Boehlau (1872)

\bibitem{Hickey}
  R.~Hickey,
  The Handbook of Language Contact,
  Chichester: Wiley-Blackwell (2010).

\bibitem{Britain}
  D.J.~Britain,
  Innovation Diffusion in Sociohistorical Linguistics,
  In book: Handbook of Historical Sociolinguistics (pp.451-464),
  Publisher: WileyEditors: J M Hernandez Campoy, J C Conde Silvestre,
  DOI:10.1002/9781118257227.ch24

\bibitem{HJ.Schmid}
  H.J.~Schmid,
  'Diffusion', The Dynamics of the Linguistic System: Usage, Conventionalization, and Entrenchment (Oxford, 2020; online edn, Oxford Academic, 20 Feb. 2020), https://doi.org/10.1093/oso/9780198814771.003.0009,
  
\bibitem{Seguy}
  J.~S\'eguy,
  La relation entre la distance spatiale et la distance lexicale,
  Revue de Linguistique Romane,
  35,138,
  (1971),
  Pages 335–357.

\bibitem{Nerbonne1}
  M.~Wieling, J.~Nerbonne, R.H.~Baayen,
  Quantitative Social Dialectology: Explaining Linguistic Variation Geographically and Socially,
  Plos One,
  Vol.6, 9,
  (2011),
  DOI:doi.org/10.1371/journal.pone.0023613

\bibitem{Chambers}
  J.~Chambers, P.~Trudgill,
  Dialectology,
  Cambridge University Press, Second edition.
  (1998)

\bibitem{Goebl}
  H.~Goebl,
  Dialectometry: A short overview of the principles and practice of
  quantitative classification of linguistic atlas data.,
  (1993) In: Kohler R, Rieger B, eds.,
  Contributions to Quantitative Linguistics. Dordrecht: Kluwer. (1984) pp 277–315.

\bibitem{EBidese}
  E.~Bidese,
  Sprachkontakt generativ,
  Volume 582 in the series Linguistische Arbeiten (2023),
  https://doi.org/10.1515/9783110765014
  
\bibitem{Nerbonne2}
  W.~Heeringa, J.~Nerbonne,
  Dialect areas and dialect continua,
  Language Variation and Change,
  (2001) 13: 375–400.

\bibitem{Kretzschmar}  W.~Kretzschmar~Jr.,
  Quantitative areal analysis of dialect features,
  Language Variation and Change,
  (1996) 8: 13–39.

\bibitem{Labov}
  W.~Labov,
  The social motivation of a sound change.,
  Word (1963) 19: 273–309
  
\bibitem{Trudgill}
  P.~Trudgill,
  Dialects in contact,
  Blackwell,
  (1986) 

\bibitem{Trudgill2}
  P.~Trudgill, 
  Linguistic Change and Diffusion: Description and Explanation in Sociolinguistic Dialect Geography,
  Language in Society,
  (1974) 2:215–246.

\bibitem{Nerbonne3}
  J.~Nerbonne,
  Measuring the diffusion of linguistic change,
  Phil. Trans. R. Soc. (2010) B3653821–3828,
  DOI: http://doi.org/10.1098/rstb.2010.0048

\bibitem{Bidese}
  E.~Bidese, A.~Tomaselli,
  Language synchronization north and south of the Brenner Pass: modeling the continuum.,
  STUF - Language Typology and Universals,
  (2021),
  74(1), 185–216.,
  DOI https://doi.org/10.1515/stuf-2021-1028

\bibitem{Tomaselli}
  A.~Tomaselli, E.~Bidese,
  Fortune and Decay of Lexical Expletives in Germanic and Romance along the Adige River.,
  Languages 8(1): 44. (2023),
  DOI: https://doi.org/10.3390/languages8010044

\bibitem{Pescarini}
  D.~Pescarini,
  Expletive Subject Clitics in Northern Italo-Romance,
  (2022) Languages 7(4): 265,
  DOI: https://doi.org/10.3390/languages7040265

\bibitem{Rabanus et al}
  S.~Rabanus, A.~Kruijt, M.~Tagliani, et al.,
  (2022) VinKo (Varieties in Contact) Corpus v1.1,
  Eurac Research CLARIN Centre,
  http://hdl.handle.net/20.500.12124/46.
  
\bibitem{folium}
  https://python-visualization.github.io/folium/index.html

\bibitem{geographicdata}
  J.~S.J.~Rey, D.~Arribas-Bel, L.J.~Wolf,
  Geographic Data Science with Python
  https://geographicdata.science/book/intro.html

\bibitem{jupyter}
  https://jupyter.org/

\bibitem{Alfeld}
  P.~Alfeld,
  A trivariate Clough-Tocher scheme for tetrahedral data.
  Computer Aided Geometric Design,
  (1984) 1, 169
  

\bibitem{Farin}
  G.~Farin,
  Triangular Bernstein-Bezier patches
  Computer Aided Geometric Design,
  (1986) 3, 83


\bibitem{scipy_interpolate}
  https://docs.scipy.org/doc/scipy/reference/generated/
  scipy.interpolate/CloughTocher2DInterpolator

\bibitem{pyplot}
  https://matplotlib.org/stable/tutorials/introductory/pyplot.html


\bibitem{Euregio}
  Euregio (a cura di),
  Tirolo Alto Adige Trentino - Uno sguardo storico,
  Trento (2013),
  ISBN 978-88-907860-2-0.

\bibitem{Zieger}
  A.~Zieger
  Storia del Trentino e dell'Alto Adige,
  (1925)

\bibitem{Obermair}
  H.~Obermair,
  Il notariato nello sviluppo della città e del suburbio di Bolzano nei secoli XII–XVI,
  in Andrea Giorgi et al. (a cura di),
  Milano, Giuffrè (2014), pp. 293-322

\bibitem{Riedmann}
  J.~ Riedmann,
  Die ältesten Aufzeichnungen in italienischer Sprache in Südtirol,
  in Der Schlern, n. 52 (1978), pp. 15-27.

\bibitem{Boltzmann}  
  L.~Boltzmann, 
  Annln Phys. 53, 959, (1894)

\bibitem{Crank}
  J.~Crank,
  The Mathematics of Diffution,
  Claredon Press second edition (1975)

\bibitem{Nervadof}
  G.~Nervadof,
  https://levelup.gitconnected.com/solving-2d-heat-equation-numerically-using-python-3334004aa01a


\bibitem{Cowan}
  G.~Cowan,
  Statistical Data Analysis,
  United Kingdom: Clarendon Press (1998)

\end{thebibliography}

\end{document}